\documentclass[10pt,twocolumn,letterpaper]{article}

\usepackage[pagenumbers]{cvpr} 
\usepackage[accsupp]{axessibility}
\usepackage{graphicx}
\usepackage{amsmath}
\usepackage{amssymb}
\usepackage{booktabs}
\usepackage{makecell}
\usepackage{array}
\usepackage{multicol}
\usepackage{wrapfig}
\usepackage{multirow}
\usepackage{bbm}
\usepackage{diagbox}
\usepackage{float}
\usepackage{graphicx}
\usepackage{perpage} 
\usepackage{enumitem}
\usepackage{algpseudocode}
\usepackage[ruled,vlined]{algorithm2e}
\usepackage{extarrows}
\usepackage{booktabs} 

\usepackage[pagebackref=true,breaklinks=true,letterpaper=true,colorlinks,urlcolor=black,bookmarks=false]{hyperref}
\hypersetup{citecolor=[RGB]{119,185,0}}

\usepackage[capitalize]{cleveref}
\crefname{section}{Sec.}{Secs.}
\Crefname{section}{Section}{Sections}
\Crefname{table}{Table}{Tables}
\crefname{table}{Tab.}{Tabs.}
\MakePerPage{footnote} 

\def\cvprPaperID{3004} 
\def\confName{CVPR}
\def\confYear{2022}

\begin{document}
\definecolor{mygray}{gray}{.92}
\definecolor{commentcolor}{RGB}{110,154,155}   
\newcommand{\PyComment}[1]{\ttfamily\textcolor{commentcolor}{\# #1}}  
\newcommand{\captionfontsize}{\fontsize{9pt}{10.8pt}\selectfont} 
\newcommand{\PyCode}[1]{\ttfamily\textcolor{blue}{#1}}
\renewcommand{\captionlabelfont}{\scriptsize}
\newcommand{\tablestyle}[2]{\setlength{\tabcolsep}{#1}
\renewcommand{\arraystretch}{#2}\centering\captionfontsize}
\hypersetup{
    urlcolor=blue,
}

\title{Panoptic SegFormer: Delving Deeper into Panoptic Segmentation with Transformers}

\author{
Zhiqi Li$^1$,
Wenhai Wang$^{2}$,
Enze Xie$^3$,
Zhiding Yu$^4$, \\
Anima Anandkumar$^{4,5}$,
Jose M. Alvarez$^4$,
Ping Luo$^3$,
Tong Lu$^{1}$
\\ [0.15cm]
$^1$Nanjing University~~$^2$Shanghai AI Laboratory~~$^3$The University of Hong Kong~~$^4$NVIDIA~~$^5$Caltech
\\ [0.15cm]
lzq@smail.nju.edu.cn~~wangwenhai@pjlab.org.cn~~xieenze@hku.hk
~~zhidingy@nvidia.com\\~~aanandkumar@nvidia.com~~josea@nvidia.com 
~~pluo@cs.hku.hk~~lutong@nju.edu.cn
}

\maketitle

\begin{abstract}
Panoptic segmentation involves a combination of joint semantic segmentation and instance segmentation, where image contents are divided into two types: things and stuff. We present Panoptic SegFormer, a general framework for panoptic segmentation with transformers. It contains three innovative components: an efficient deeply-supervised mask decoder, a query decoupling strategy, and an improved post-processing method. We also use Deformable DETR to efficiently process multi-scale features, which is a fast and efficient version of DETR. Specifically, we supervise the attention modules in the mask decoder in a layer-wise manner. This deep supervision strategy lets the attention modules quickly focus on meaningful semantic regions. It improves performance and reduces the number of required training epochs by half compared to Deformable DETR. Our query decoupling strategy decouples the responsibilities of the query set and avoids mutual interference between things and stuff. In addition, our post-processing strategy improves performance without additional costs by jointly considering classification and segmentation qualities to resolve conflicting mask overlaps. Our approach increases the accuracy 6.2\% PQ over the baseline DETR model. Panoptic SegFormer achieves state-of-the-art results on COCO test-dev with 56.2\% PQ. It also shows stronger zero-shot robustness over existing methods.  The code  is released at \url{https://github.com/zhiqi-li/Panoptic-SegFormer}.

\end{abstract}
\vspace{-0.1in}
\section{Introduction}
\label{sec:intro}

\begin{figure}[t]
	\renewcommand{\captionlabelfont}{\captionfontsize}
	\centering
	\includegraphics[width=1\columnwidth]{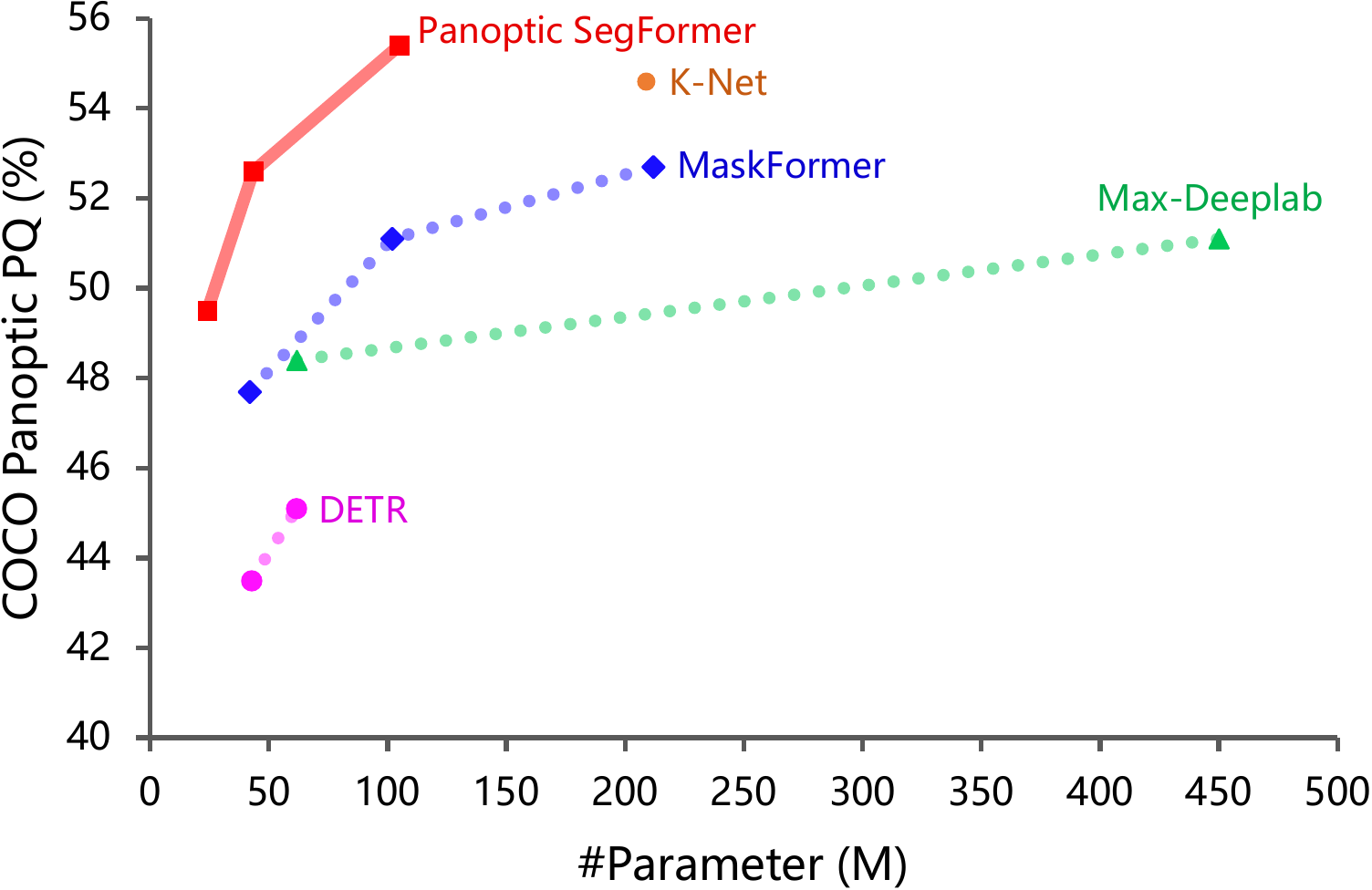}
	\hspace{-50mm}\resizebox{0.56\columnwidth}{!}{\tablestyle{2pt}{1}
	\renewcommand\arraystretch{0.9}
	\begin{tabular}[b]{l|c|c}
	 & PQ (\%)&\#Param (M)  \\
	\hline
	DETR-R50~\cite{carion2020end} & 43.5 & 42.8\\
    Max-Deeplab-S~\cite{wang2021max} & 48.4&62.0 \\
    Max-Deeplab-L~\cite{wang2021max} & 51.1& 451.0 \\
    MaskFormer-T~\cite{cheng2021per} & 47.4 & 42.0 \\
    MaskFormer-B~\cite{cheng2021per} & 51.1 & 102.0 \\
    MaskFormer-L~\cite{cheng2021per} & 52.7 & 212.0 \\
    K-Net-L~\cite{zhang2021k} & 54.6 & 208.9 \\
    \hline
    Panoptic SegFormer-B0 & 49.5 & 24.2\\
    Panoptic SegFormer-B2 & 52.5 & 43.6\\
	Panoptic SegFormer-B5 & 55.4 & 104.9\\
			\multicolumn{3}{l}{
			\vspace{9mm}}\\
\end{tabular}
	}
   \vspace{-3mm}
   \caption{\captionfontsize {Comparison to the prior arts in panoptic segmentation methods on the COCO val2017 split.}
    Panoptic SegFormer models outperform the other counterparts among different models. Panoptic SegFormer (PVTv2-B5~\cite{wang2021pvtv2}) achieves 55.4\% PQ, surpassing previous methods with significantly fewer parameters.
    } 
    \vspace{-0.20in}
\label{fig:params}
\end{figure}

Semantic segmentation and instance segmentation are two important and related vision tasks. Their underlying connections recently motivated panoptic segmentation as a unification of both the tasks~\cite{kirillov2019panoptic}. In panoptic segmentation, image contents are divided
into two types: things and stuff. Things refer to countable instances (\eg, person, car) and each instance has a unique id to distinguish it from the other instances. Stuff refers to the amorphous and uncountable regions (\eg, sky, grassland) and has no instance id~\cite{kirillov2019panoptic}.

\begin{figure*}[t]
\renewcommand{\captionlabelfont}{\captionfontsize}
\begin{center}
\includegraphics[width=0.99\linewidth]{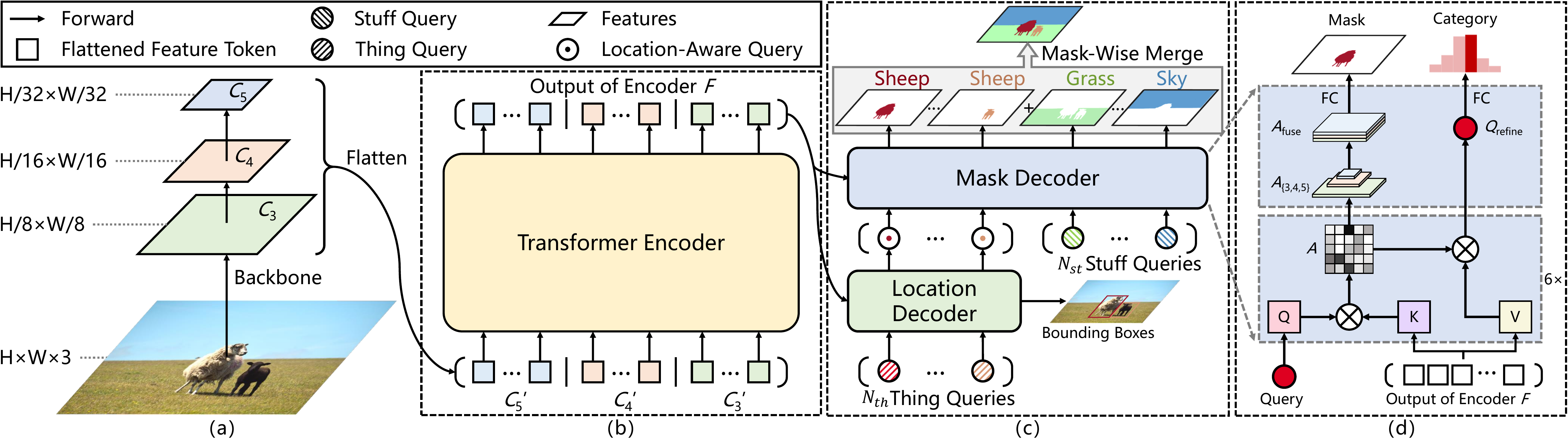}
\end{center}
\vspace{-0.15in}
\caption{\captionfontsize
\textbf{Overview of Panoptic SegFormer.} Panoptic SegFormer is composed of backbone, encoder, and decoder. The backbone and the encoder output and refine multi-scale features. Inputs of the location decoder are $N_{\rm th}$ thing queries and the multi-scale features. We feed $N_{\rm th}$ thing queries from the location decoder and $N_{\rm st}$ stuff queries to the mask decoder. The location decoder aims to learn reference points of queries, and the mask decoder predicts the final category and mask. Details of the decoder will be introduced below. We use a mask-wise merging method instead of the commonly used pixel-wise argmax method to perform inference.}
\label{fig:model}
\vspace{-0.2in}
\end{figure*}

Recent works~\cite{carion2020end,cheng2021per,wang2021max} attempt to employ transformers to handle both things and stuff through a query set. For example, DETR~\cite{carion2020end} simplifies the workflow of panoptic segmentation by adding a panoptic head on top of an end-to-end object detector.
Unlike previous methods~\cite{kirillov2019panopticfpn,kirillov2019panoptic}, DETR does not require additional handcrafted pipelines~\cite{bodla2017soft,xiong2019upsnet}. 
While being simple, DETR also causes some issues: (1) It requires a
lengthy training process to converge; (2) Because the computational complexity of self-attention is squared with the length of the input sequence, the feature resolution of DETR is limited.
So that it uses an FPN-style~\cite{lin2017feature,carion2020end} panoptic head to generate masks, which always suffer low-fidelity boundaries; (3) It handles things and stuff equally, yet
representing them with bounding boxes, which may be suboptimal for stuff~\cite{cheng2021per,wang2021max}. 
Although DETR achieves excellent performance on the object detection task, its superiority on panoptic segmentation has not been well demonstrated. In order to overcome the defects of DETR on panoptic segmentation, we propose a series of novel and effective strategies that improve the performance of transformer-based panoptic segmentation models by a large margin.

\textbf{Our approach.} In this work, we propose Panoptic SegFormer, a concise and effective framework for panoptic segmentation with transformers. Our framework design is  motivated by the following observations: 1) Deep supervision matters in learning high-qualities discriminative attention representations in the mask decoder. 2) Treating things and stuff with the same recipe~\cite{carion2020end} is suboptimal due to the different properties between things and stuff~\cite{kirillov2019panoptic}. 3) Commonly used post-processing such as pixel-wise argmax~\cite{carion2020end,cheng2021per,wang2021max} tends to generate false-positive results due to extreme anomalies. We overcome these challenges in Panoptic SegFormer framework as follows:
\begin{itemize}[leftmargin=*]
\vspace{-2mm}
\item
We propose a mask decoder that utilizes multi-scale attention maps to generate high-fidelity masks. The mask decoder is deeply-supervised, promoting discriminative attention representations in the intermediate layers with better mask qualities and faster convergence.
\vspace{-2mm}

\item 
We propose a query decoupling strategy that decomposes the query set into a thing query set to match things via bipartite matching and another stuff query set to process stuff with class-fixed assign. This strategy avoids mutual interference between things and stuff within each query and significantly improves the qualities of stuff segmentation. Kindly refer to \cref{qds} and \cref{fig:decoupling} for more details. 
\vspace{-6mm}

\item 
We propose an improved post-processing method to generate results in panoptic format. Besides being more efficient than the widely used pixel-wise argmax method, our method contains a mask-wise merging strategy that considers both classification probability and predicted mask qualities. Our post-processing method alone renders a 1.3\% PQ improvement to DETR~\cite{carion2020end}.
\vspace{-1mm}
\end{itemize}

We conduct extensive experiments on COCO~\cite{lin2014microsoft} dataset.
As shown in ~\cref{fig:params}, Panoptic SegFormer significantly surpasses priors arts such as MaskFormer~\cite{cheng2021per} and K-Net~\cite{zhang2021k} with much fewer parameters. With deformable attention~\cite{zhu2020deformable} and our deeply-supervised mask decoder, our method requires much fewer training epochs than previous transformer-based methods (24 \vs 300$+$)~\cite{carion2020end,cheng2021per}. In addition, our approach also achieves competitive performance with current methods~\cite{fang2021queryinst,chen2019hybrid} on the instance segmentation task.

\vspace{-1mm}
\section{Related Work}

\textbf{Panoptic Segmentation.}
Panoptic segmentation becomes a popular task for holistic scene understanding~\cite{kirillov2019panoptic, bonde2020towards,li2020unifying ,cheng2020panoptic}.
The panoptic segmentation literature mainly treats this problem as a joint task of instance segmentation and semantic segmentation where things and stuff are handled separately~\cite{yang2019deeperlab,gao2019ssap}. Kirillov \etal~\cite{kirillov2019panoptic} proposed the concept of and benchmark of panoptic segmentation together with a baseline that directly combines the outputs of individual instance segmentation and semantic segmentation models. Since then, models such as Panoptic FPN~\cite{kirillov2019panopticfpn}, UPSNet~\cite{xiong2019upsnet} and AUNet~\cite{li2019attention} have improved the accuracy and reduced the computational overhead by combining instance segmentation and semantic segmentation into a single model. However, these methods approximate the target task by solving the surrogate sub-tasks, therefore introducing undesired model complexities and suboptimal performance.

Recently, efforts have been made to unify the framework of panoptic segmentation.
Li \etal~\cite{li2021fully} proposed Panoptic FCN where the panoptic segmentation pipeline is simplified with a ``top-down meets bottom-up'' two-branch design similar to CondInst~\cite{tian2020conditional}. In their work, things and stuff are jointly modeled by an object/region-level kernel branch and an image-level feature branch. Several recent works represent things and stuff as queries and perform end-to-end panoptic segmentation via transformers.
DETR~\cite{carion2020end} predicts the bounding boxes of things and stuff and combines the attention maps of the transformer decoder and the feature maps of ResNet~\cite{he2016deep} to perform panoptic segmentation.
Max-Deeplab~\cite{wang2021max} directly predicts object categories and masks through a dual-path transformer regardless of the category being things or stuff. 
On top of DETR, MaskFomer~\cite{cheng2021per} used an additional pixel decoder to refine high spatial resolution features and generated the masks by multiplying queries and features from the pixel decoder.
Due to the computational complexity of self attention~\cite{vaswani2017attention},
both DETR and MaskFormer use feature maps with limited spatial resolutions for panoptic segmentation, which hurts the performance and requires combining additional high-resolution feature maps in final mask prediction.
Unlike the methods mentioned above, our query decoupling strategy deals with things and stuff with separate query sets. 
Although thing and stuff queries are designed for different targets, they are processed by the mask decoder with the same workflow. Prediction results of these queries are in the same format so that we can process them in an equal manner during the post-processing procedure. One concurrent work~\cite{zhang2021k} employs a similar line of thinking to use dynamic kernels to perform instance and semantic segmentation, and it aims to utilize unified kernels to handle various segmentation tasks. In contrast to it, we aim to delve deeper into the transformer-based panoptic segmentation. Due to the different nature of various tasks, whether a unified pipeline is suitable for these tasks is still an open problem. In this work, we utilize an additional location decoder to assist things to learn location clues and get better results.

\textbf{End-to-end Object Detection.} The recent popular end-to-end object detection frameworks have inspired many other related works~\cite{dong2021solq,fang2021queryinst}. DETR~\cite{carion2020end} is arguably the most representative end-to-end object detector among these methods. DETR models the object detection task as a dictionary lookup problem with learnable queries and employs an encoder-decoder transformer to predict bounding boxes without extra post-processing. DETR greatly simplifies the conventional detection framework and removes many hand-crafted components such as Non-Maximum Suppression (NMS)~\cite{ren2015faster,lin2017focal} and anchors~\cite{lin2017focal}. Zhu \etal~\cite{zhu2020deformable} proposed Deformable DETR, which further reduces the memory and computational cost through deformable attention layers. In this work, we adopt deformable attention~\cite{zhu2020deformable} for the improved efficiency and convergence over DETR~\cite{carion2020end}.

\vspace{-2mm}
\section{Methods}
\subsection{Overall Architecture}

As illustrated in ~\cref{fig:model}, 
Panoptic SegFormer consists of three key modules: transformer encoder, location decoder, and mask decoder, where (1) the transformer encoder is applied to refine the multi-scale feature maps given by the backbone, (2) the location decoder is designed to capturing location clues of things, and (3) the mask decoder is for final classification and segmentation.

Our architecture feeds an input image $X\!\in\!\mathbb{R}^{H \times W\times 3}$ to the backbone network, and obtains the feature maps $C_3$, $C_4$, and $C_5$ from the last three stages, of which the resolutions are 
$1/8$, $1/16$ and $1/32$ compared to the input image, respectively.
We project the three feature maps to the ones with 256 channels by a fully-connected (FC) layer, 
and flatten them into feature tokens $C'_3$, $C'_4$, and $C'_5$.
Here, we define $L_i$ as $\frac{H}{2^{i+2}}\!\times\!\frac{W}{2^{i+2}}$, and the shapes of $C'_3$, $C'_4$, and $C'_5$ are $L_1\!\times\!256$, $L_2\!\times\!256$, and $L_3\!\times\!256$, respectively.
Next, using the concatenated feature tokens as input,
the transformer encoder outputs the refined features of size $(L_1\!+\!L_2\!+\!L_3)\!\times\!256$.
After that, we use $N_{\rm th}$ and $N_{\rm st}$ randomly initialized things and stuff queries to describe things and stuff separately. Location decoder refines $N_{\rm th}$ thing queries by detecting the bounding boxes of things to capture location information. The mask decoder then takes both things and stuff queries as input and predicts mask and category at each layer. 

During inference, we adopt a mask-wise merging strategy 
to convert the predicted masks from final mask decoder layer into the panoptic segmentation results, which will be introduced in detail in ~\cref{mask-wise}.
\begin{figure}[t]
\renewcommand{\captionlabelfont}{\captionfontsize}
\begin{center}
   \includegraphics[width=\linewidth]{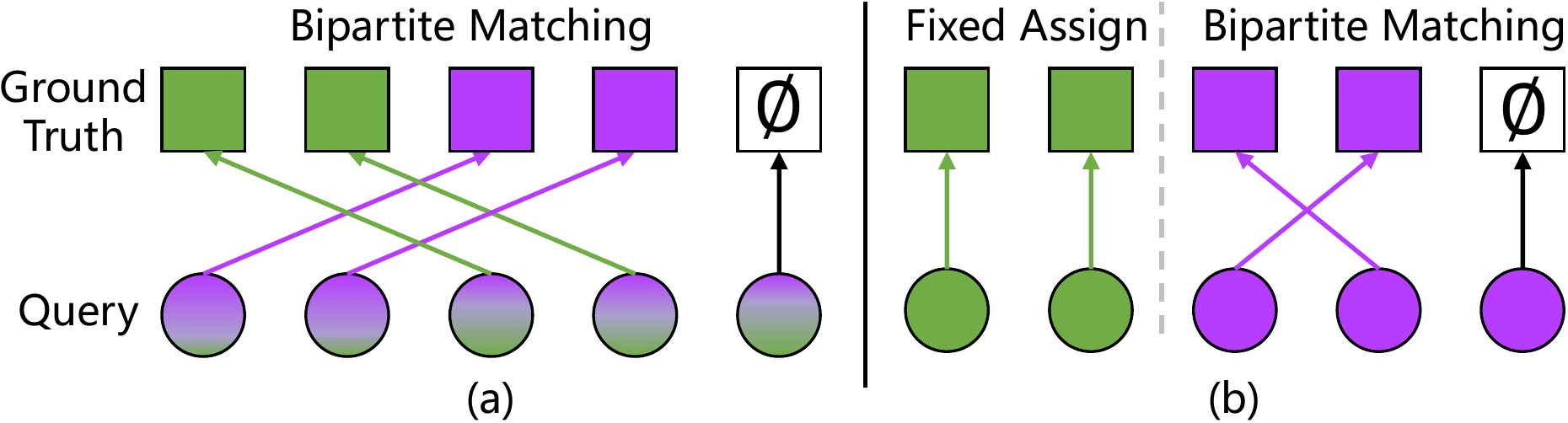}
\end{center} 
\vspace{-0.20in}  
   \caption{ \captionfontsize (a)  Methods~\cite{carion2020end,wang2021max,cheng2021per} adopt one query set to match things (purple squares) and stuff (green squares) jointly. (b) We use one thing query set (purple circles) to target things through bipartite matching and one stuff query set (green circles) to predict stuff by a class-fixed assign strategy. $\varnothing$ is assigned to not-matched queries.
    }
\label{fig:decoupling}
\vspace{-0.20in} 
\end{figure}
\vspace{-1mm}
\subsection{Transformer Encoder}
\vspace{-1mm}
High-resolution and the multi-scale features maps are
important for the segmentation tasks~\cite{kirillov2019panopticfpn,wang2020solov2,li2021fully}.
Since the high computational cost of self-attention layer,
previous transformer-based methods~\cite{carion2020end,cheng2021per} can only process low-resolution feature maps 
(\eg,  ResNet $C_5$) in their encoders, which limits the segmentation performance.
Different from these methods, we employ the deformable attention~\cite{zhu2020deformable} to implement our transformer encoder. Due to the low computational complexity of the deformable attention,
our encoder can refine and involve positional encoding~\cite{vaswani2017attention} to high-resolution and multi-scale feature maps $F$.
\vspace{-1mm}
\subsection{Decoder}
\vspace{-1mm}
In this section, we introduce our query decoupling strategy firstly, and then we will explain the details of our location decoder and mask decoder.
\vspace{-4mm}
\subsubsection{Query Decoupling Strategy}\label{qds}
\vspace{-1mm}
We argue that using one query set to handle both things and stuff equally is suboptimal. Since there many different properties between them, things and stuff is likely to interfere with each other and hurt the model performance, especially for PQ$^{\rm st}$.
To prevent things and stuff from interfering with each other, we apply a query decoupling strategy in Panoptic SegFormer, as shown in \cref{fig:decoupling}. Specifically, $N_{\rm th}$ thing queries are used to predict things results, and $N_{\rm st}$ stuff queries target stuff only. Using this form, things and stuff queries can share the same pipeline since they are in the same format. We can also customize private workflow for things or stuff according to the characteristics of different tasks.
In this work, we use an additional location decoder to detect individual instances with thing queries, and this will assist  in distinguishing between different instances~\cite{kirillov2019panoptic}. Mask decoder accepts both thing queries and stuff queries and generates the final masks and categories. Note that, for thing queries, ground truths are assigned by bipartite matching strategy. For stuff, We use a class-fixed assign strategy, and each stuff query corresponds to one stuff category. 

Thing and stuff queries will output results in the same format, and we handle these results with a uniform post-processing method.
\vspace{-4mm}
\subsubsection{Location Decoder}
\vspace{-1mm}
Location information plays an important role in distinguishing things with different instance ids in the panoptic segmentation task~\cite{wang2020solo,tian2020conditional,wang2020solov2}.
Inspired by this, we employ a location decoder to introduce the location information of things into the learnable queries. 
Specifically, given $N_{\rm th}$ randomly initialized thing queries and the refined feature tokens generated by transformer encoder, the decoder will output $N_{\rm th}$ location-aware queries.

In the training phase, we apply an auxiliary MLP head on top of location-aware queries to predict the bounding boxes and categories of the target object,
We supervise the prediction results with a detection loss $\mathcal{L}_{\rm det}$. The MLP head is an auxiliary branch, which can be discarded during the inference phase.
The location decoder follows Deformable DETR~\cite{zhu2020deformable}. Notably, the location decoder can learn location information by predicting the mass centers of masks instead of bounding boxes.
This box-free model can still achieve comparable results to our box-based model.

\vspace{-4mm}
\subsubsection{Mask Decoder}
\vspace{-1mm}

As shown in ~\cref{fig:model}~(d), 
the mask decoder is proposed to predict the categories and masks according to the given queries.
The queries $Q$ of the mask decoder are the location-aware thing queries from the location decoder or the class-fixed stuff queries. The keys $K$ and values $V$ of the mask decoder are projected from the refined feature tokens $F$ from the transformer encoder.
We first pass thing queries through the mask decoder, and then fetch the attention map $A\in \mathbb{R}^{N \times h \times (L_1+L_2+L_3)}$ and the refined query $Q_{\rm refine} \in \mathbb{R}^{N\times 256}$ from  each decoder layer, where $N\!=\!N_{\rm th}\!+\!N_{\rm st}$ is the whole query number, $h$ is the number of attention heads, and $L_1\!+\!L_2\!+\!L_3$ is the length of feature tokens $F$.

Similar to methods~\cite{wang2021max,carion2020end}, we directly perform classification through a FC layer on top of the refined query $Q_{\rm refine}$ from each decoder layer.  Each thing query needs to predict probabilities over all thing categories. Stuff query only predicts the probability of its corresponding stuff category.

At the same time, to predict the masks, we first split and reshape the attention maps $A$ into 
attention maps $A_3$, $A_4$, and $A_5$, which have the same spatial resolution as $C_3$, $C_4$, and $C_5$.
This process can be formulated as:
\vspace{-1.5mm}
\begin{equation}\label{split}
(A_3, A_4, A_5) = {\rm Split}(A),\ \ \ A_{i}\in\mathbb{R}^{ \frac{H}{2^{i+2}}\!\times\!\frac{W}{2^{i+2}}\!\times h },
\vspace{-1mm}
\end{equation}
where ${\rm Split}(\cdot)$ denotes the split and reshaping operation. After that, as illustrated in ~\cref{fused}, we upsample these attention maps to the resolution of $H/8\!\times\! W/8$ and concatenate them along the channel dimension,
\vspace{-1.5mm}
\begin{equation}\label{fused}
A_{\rm fused} = {\rm Concat}(A_1, {\rm Up}_{\times 2}(A_2),{\rm Up}_{\rm \times 4}(A_3)),
\vspace{-1mm}
\end{equation}
\noindent where ${\rm Up}_{\times 2}(\cdot)$ and ${\rm Up}_{\times 4}(\cdot)$ mean the 2 times and 4 times bilinear interpolation operations, respectively. ${\rm Concat}(\cdot)$ is the concatenation operation.
Finally, based on the fused attention maps $A_{\rm fused}$, we predict the binary mask through a $1\!\times\!1$ convolution.

Previous literature~\cite{zhu2020deformable} argues that the reason for slow convergence of DETR is that attention modules equally pay attention to all the pixels in the feature maps, and learning to focus on sparse meaningful locations requires plenty of effort. We use two key designs to solve this problem in our mask decoder: (1) Using an ultra-light FC head to generate masks from the attention maps, ensuring  attention modules can be guided by ground truth mask to learn where to focus on. This FC head only contains $200$ parameters, which ensures the semantic information of attention maps is highly related to the mask. Intuitively, the ground truth mask is exactly the meaningful region on which we expect the attention module to focus.
(2) We employ deep supervision in the mask decoder. Attention maps of each layer will be supervised by the mask, the attention module can capture meaningful information in the earlier stage. This can highly accelerate the learning process of attention modules.

\subsection{Loss Function}
During training, our overall loss function of Panoptic SegFormer can be written as:

\vspace{-3mm}
\begin{equation}\label{loss}
\mathcal{L} = 
\lambda_{\rm things}\mathcal{L}_{\rm things} + \lambda_{\rm stuff}\mathcal{L}_{\rm stuff},
\vspace{-1mm}
\end{equation}
where ${L}_{\rm things}$ and ${L}_{\rm stuff}$ are loss for things and stuff, separately. $\lambda_{\rm things}$ and $\lambda_{\rm stuff}$ are hyperparameters.

\textbf{Things Loss.} Following common practices~\cite{carion2020end,stewart2016end}, we search the best  bipartite matching between the prediction set and the ground truth set.
Specifically, we utilize Hungarian algorithm~\cite{kuhn1955hungarian} to search for the permutation with the minimum matching cost, which is the sum of the classification loss $\mathcal{L}_{\rm cls}$, detection loss $\mathcal{L}_{\rm det}$ and the segmentation loss $\mathcal{L}_{\rm seg}$.
The overall loss function for the thing categories is accordingly defined as follows:
\vspace{-3mm}
\begin{equation}\label{things_loss}
\mathcal{L}_{\rm things} =\lambda_{\rm det}\mathcal{L}_{\rm det}+ \sum_{i}^{D_m}{(
\lambda_{\rm cls}\mathcal{L}_{\rm cls}^{i} + \lambda_{\rm seg}\mathcal{L}_{\rm seg}^{i})},
\vspace{-3mm}
\end{equation}
where $\lambda_{\rm cls}$, $\lambda_{\rm seg}$, and $\lambda_{\rm loc}$ are the weights to balance three losses. $D_m$ is the number of layers in the  mask decoder.
$\mathcal{L}_{\rm cls}^{i}$ is the classification loss that is implemented by Focal loss~\cite{lin2017focal}, and $\mathcal{L}_{\rm seg}^{i}$ is the segmentation loss implemented by Dice loss~\cite{milletari2016v}. $\mathcal{L}_{\rm det}$ is the loss of Deformable DETR that used to perform detection.

\textbf{Stuff Loss.} We use a fixed matching strategy for stuff. Thus there is a one-to-one mapping between stuff queries and stuff categories. The loss for the stuff categories is similarly defined as: 
\vspace{-3mm}
\begin{equation}\label{stuff_loss}
\mathcal{L}_{\rm stuff} =\sum_{i}^{D_m}{(
\lambda_{\rm cls}\mathcal{L}_{\rm cls}^{i} + \lambda_{\rm seg}\mathcal{L}_{\rm seg}^{i})}
,
\vspace{-3mm}
\end{equation}
where $\mathcal{L}_{\rm cls}^{i}$ and $\mathcal{L}_{\rm seg}^{i}$ are the same as those in \cref{things_loss}.

\subsection{Mask-Wise Merging Inference}\label{mask-wise}
Panoptic Segmentation requires each pixel to be assigned a category label (or void) and instance id (ignored for stuff)~\cite{kirillov2019panoptic}.
One challenge of panoptic segmentation is that it requires generating non-overlap results. Recent methods~\cite{cheng2021per,carion2020end,wang2021max} directly use pixel-wise argmax to determine the attribution of each pixel, and this can solve the overlap problem naturally. Although pixel-wise argmax strategy is simple and effective, we observe that it consistently produces false-positive results due to the abnormal pixel values.

Unlike pixel-wise argmax resolves conflicts on each pixel, we propose the mask-wise merging strategy by resolving the conflicts among predicted masks. Specifically, we use the confidence scores of masks to determine the attribution of the overlap region. Inspired by previous NMS methods~\cite{wang2020solov2}, the confidence scores take into both classification probability and predicted mask qualities. The confidence score of the i-th result can be formulated as:
\vspace{-2mm} 
\begin{equation}\label{confidence}
s_i\!=\!p_i^\alpha\!\times\!{\rm average}({\mathbbm{1}_{\{m_i[h,w]>0.5\}}m_i[h,w])}^\beta,
\vspace{-2mm} 
\end{equation}
where $p_i$ is the most likely class probability of i-th result. $m_i[h,w]$ is the mask logit at pixel $[h,w]$, $\alpha,\beta$ are used to balance the weight of classification probability and segmentation qualities.

\begin{algorithm}[t]
\SetAlgoLined
\PyCode{\small{def}} \PyCode{\small{MaskWiseMergeing}}\small{(c,s,m):}\\
    \Indp
        \PyComment{\small{category ${\rm c}\in\mathbb{R}^N$}} \\
        \PyComment{\small{confidence score ${\rm s}\in\mathbb{R}^N$}} \\
        \PyComment{\small{mask  ${\rm m}\in\mathbb{R}^{N\!\times\!H\!\times\!W}$ }}
        
        \small{SemMsk = np.zeros(H,W)}\\
        \small{IdMsk = np.zeros(H,W)}\\
        \small{order = np.argsort(-s)}\\
        \small{id = 0}\\
        \PyCode{\small{for}} \small{i} \PyCode{\small{in}} \small{order:} \\
        \Indp   
            \small{m$_i$ = m[i]>0.5 \& (SemMsk==0)}\\
            \PyCode{\small{if}} \small{s[i]< t$_{\rm cnf}$} \PyCode{\small{or}} \small{$\frac{m_{i}}{m[i]>0.5}$ < t$_{\rm keep}$:}\\
            \Indp
                \PyCode{\small{continue}}\\
            \Indm
            \small{SemMsk[m$_i$] = c[i]}\\
            \small{IdMsk[m$_i$] = id}\\
            \small{id += 1}\\
            \Indm
        \Indm 
        \PyCode{\small{return}} \small{SemMsk,IdMsk}\\
    \Indm 
\caption{\small{\textbf{Mask-Wise Merging}}}
\label{inference}
\end{algorithm}

As illustrated in ~\cref{inference},
mask-wise merging strategy takes $c$, $s$, and $m$ as input, denoting the predicted categories, confidence scores, and segmentation masks, respectively. It outputs a semantic mask $\tt SemMsk$ and an instance 
id mask $\tt IdMsk$, to assign a category label and an instance id to each pixel.
Specifically,  $\tt SemMsk$  and $\tt IdMsk$ are first initialized by zeros. 
Then, we sort prediction results in descending order of confidence score and fill the sorted predicted masks into $\tt SemMsk$  and $\tt IdMsk$ in order. Then we discard the results with confidence scores below ${\rm t}_{\rm cls}$ and remove the overlaps with lower confidence scores.
Only remained non-overlap part with a sufficient fraction ${\rm t}_{\rm keep}$ to origin mask will be kept.
Finally, the category label and unique id of each mask are added to generate non-overlap panoptic format results.
\vspace{-1mm}
\section{Experiments}
\vspace{-1mm} 
We evaluate Panoptic SegFormer on COCO~\cite{lin2014microsoft} and ADE20K dataset~\cite{zhou2017scene}, comparing it with several state-of-the-art methods. We provide the main results of panoptic segmentation and instance segmentation. We also conduct detailed ablation studies to verify the effects of each module. Please refer to Appendix for implementation details.
\vspace{-1mm} 
\subsection{Dataset}
\vspace{-1mm} 
We perform experiments on COCO 2017 datasets~\cite{lin2014microsoft} without external data. The COCO dataset contains 118K training images and 5k validation images, and it contains 80 things and 53 stuff. We further demonstrate the generality of our model on the ADE20K dataset~\cite{zhou2017scene}, which contains 100 things and 50 stuff.
\vspace{-1mm}  
\subsection{Main Results}
\vspace{-1mm} 
\textbf{Panoptic segmentation.} We conduct experiments on COCO \texttt{val} set and \texttt{test-dev} set. In ~\cref{val} and ~\cref{test_dev}, we report our main results, comparing with other state-of-the-art methods. Panoptic SegFormer attains 49.6\% PQ on COCO \texttt{val} with ResNet-50 as the backbone and single-scale input, and it surpasses previous methods K-Net~\cite{zhang2021k} and DETR~\cite{carion2020end} over 2.5\% PQ and 6.2\% PQ, respectively. Except for the remarkable performance, the training of Panoptic SegFormer is efficient. Under $1\!\times$ training strategy (12 epochs), Panoptic SegFormer (R50) achieves 48.0\% PQ that outperforms MaskFormer\cite{cheng2021per} that training 300 epochs by 1.5\% PQ. 
Enhanced by vision transformer backbone Swin-L~\cite{liu2021swin}, Panoptic SegFormer attains a new record of 56.2\% PQ on COCO \texttt{test-dev} without bells and whistles, surpassing  MaskFormer~\cite{cheng2021per} over 2.9\% PQ. 
Our method even surpasses the previous competition-level method Innovation~\cite{chen56joint} over 2.7 \% PQ. We also obtain comparable performance by employing PVTv2-B5~\cite{wang2021pvtv2}, while the model parameters and FLOPs are reduced significantly compared to Swin-L. Panoptic SegFormer also outperforms MaskFormer by 1.7\% PQ on ADE20K dataset~\cite{zhou2017scene}, see \cref{ade}.

\begin{table}[t]
\renewcommand{\captionlabelfont}{\captionfontsize}
\renewcommand\arraystretch{0.9}
\begin{center}

\resizebox{\linewidth}{!}{
\addtolength{\tabcolsep}{-3pt}
\begin{tabular}{l c c c c cc c}
\toprule
Method& Backbone  & Epochs & PQ & ${\rm PQ}^{\rm th}$ & ${\rm PQ}^{\rm st}$ & \#P & \#F \\
\midrule

Panoptic FPN~\cite{kirillov2019panopticfpn} & R50 & 36& 41.5&48.5&31.1 &-&- \\
SOLOv2~\cite{wang2020solov2} & R50 &   36 & 42.1 & 49.6 & 30.7&-&-   \\
DETR~\cite{carion2020end} & R50 & 325 & 43.4 & 48.2 & 36.3 &42.9&248 \\
Panoptic FCN~\cite{li2021fully} & R50 &  36 & 43.6 & 49.3 & 35.0 &  37.0&244\\
K-Net~\cite{zhang2021k} & R50 & 36 & 47.1 & 51.7 & 40.3  &-&- \\
MaskFormer~\cite{cheng2021per} & R50& 300 & 46.5& 51.0& 39.8 & 45.0 & 181\\
Panoptic SegFormer &  R50 &   \textbf{12} & 48.0 & 52.3 & 41.5 &51.0 & 214\\
Panoptic SegFormer&  R50 &   24 & \textbf{49.6} & \textbf{54.4} & \textbf{42.4} & 51.0 & 214 \\
\midrule
DETR~\cite{carion2020end} & R101 &    325 & 45.1 & 50.5 & 37.0 &61.8& 306\\
Max-Deeplab-S~\cite{wang2021max} & Max-S~\cite{wang2021max}  &   54 & 48.4 & 53.0 & 41.5 &61.9 &162 \\
MaskFormer~\cite{cheng2021per}  & R101& 300 & 47.6& 52.5& 40.3 & 64.0 & 248\\
Panoptic SegFormer&  R101 &  24 &\textbf{50.6} & \textbf{55.5}  & \textbf{43.2}  &69.9 &286\\
\midrule
Max-Deeplab-L~\cite{wang2021max} & Max-L~\cite{wang2021max}  &   54 & 51.1 & 57.0 & 42.2 &451.0 &1846 \\
Panoptic FCN~\cite{li2021fully2} & Swin-L{$^{\text{\textdagger}}$} & 36 & 51.8 & 58.6&41.6& -&-\\
MaskFormer~\cite{cheng2021per} &  Swin-L{$^{\text{\textdagger}}$}& 300 & 52.7 & 58.5 & 44.0 & 212.0 &792\\

K-Net~\cite{zhang2021k} & Swin-L{$^{\text{\textdagger}}$}& 36 & 54.6 & 60.2 & 46.0 & 208.9 & - \\
Panoptic SegFormer&  Swin-L{$^{\text{\textdagger}}$} &  24 &\textbf{55.8} & \textbf{61.7}  & \textbf{46.9}  &221.4 &816\\
Panoptic SegFormer &  PVTv2-B5{$^{\text{\textdagger}}$}  &   24   &{55.4} & 61.2  & 46.6  &104.9 &349\\
\bottomrule
\end{tabular}
}
\end{center}
\vspace{-0.20in}  
\caption{ \captionfontsize
{Experiments on COCO val set.} \#P and \#F indicate number of parameters (M) and number of FLOPs (G). Panoptic SegFormer (R50) achieves 49.6\% PQ on COCO \texttt{val}, surpassing previous methods such as DETR~\cite{carion2020end} and MaskFormer~\cite{cheng2021per} over 6.2\% PQ and 3.1\% PQ respectively.
$^{\text{\textdagger}}$ notes that backbones are pre-trained on ImageNet-22K.}
\label{val}
\vspace{-0.10in}  
\end{table}

\begin{table}[t]
\renewcommand{\captionlabelfont}{\captionfontsize}
\renewcommand\arraystretch{0.9}
\begin{center}
\renewcommand\tabcolsep{5pt} 

\resizebox{\linewidth}{!}{
\begin{tabular}{l c c c c c c }
\toprule
Method& Backbone & PQ & PQ$^{\rm th}$ & PQ$^{\rm st}$ & SQ & RQ\\

\midrule
Max-Deeplab-L~\cite{wang2021max} & Max-L~\cite{wang2021max} & 51.3 & 57.2 & 42.4  & 82.5 & 61.3 \\
Innovation~\cite{chen56joint} & ensemble  & 53.5 & {61.8} & 41.1 &\textbf{83.4}&63.3 \\
MaskFormer~\cite{cheng2021per} &  Swin-L{$^{\text{\textdagger}}$}& 53.3 & 59.1& 44.5 & 82.0& 64.1 \\
K-Net~\cite{zhang2021k} &  Swin-L{$^{\text{\textdagger}}$} & 55.2 &  61.2 & 46.2 & 82.4  &66.1 \\
\midrule
Panoptic SegFormer  &  R50 & 50.2 & 55.3 & 42.4 &81.9&60.4 \\
Panoptic SegFormer &  R101 & {50.9} &{56.2}& 43.0& 82.0&61.2  \\
Panoptic SegFormer &  Swin-L{$^{\text{\textdagger}}$} & \textbf{56.2} &\textbf{62.3}& \textbf{47.0} & 82.8 & \textbf{67.1}   \\
Panoptic SegFormer &  PVTv2-B5{$^{\text{\textdagger}}$} & {55.8} &{61.9}&{46.5} & 83.0 & {66.5}   \\
\bottomrule
\end{tabular}}
\end{center}
\vspace{-0.20in}  
\caption{ \captionfontsize
{Experiments on COCO test-dev set.}  $^{\text{\textdagger}}$ notes that backbones are pre-trained on ImageNet-22K.
}
\vspace{-0.1in}
\label{test_dev}
\end{table}

\begin{table}[t]

\renewcommand{\captionlabelfont}{\captionfontsize}
\renewcommand\tabcolsep{4pt} 
\renewcommand\arraystretch{0.9}
\begin{center}

\resizebox{\linewidth}{!}{

\begin{tabular}{l c c c c c c }
\toprule
Method& Backbone & PQ & PQ$^{\rm th}$ & PQ$^{\rm st}$ & SQ & RQ\\

\midrule
BGRNet~\cite{wu2020bidirectional}& R50 & 31.8 & - & -  & - & - \\
Auto-Panoptic~\cite{wu2020auto} & ShuffleNetV2~\cite{ma2018shufflenet}  & 32.4 &- & - &-&- \\
MaskFormer~\cite{cheng2021per} & R50 & 34.7 & 32.2& \textbf{39.7} & 76.7& 42.8 \\
MaskFormer~\cite{cheng2021per} & R101 & 35.7 & 34.5& 38.0 & {77.4}& 43.8 \\
\midrule
Panoptic SegFormer  &  R50 & \textbf{36.4} & \textbf{35.3} & 38.6 & \textbf{78.0}& \textbf{44.9}\\
\bottomrule
\end{tabular}
}
\end{center}
\vspace{-0.20in}  
\caption{ \captionfontsize
{Panoptic segmentation results on ADE20K val set.} 
}
\label{ade}
\vspace{-0.20in}  
\end{table}

\textbf{Instance segmentation.} Panoptic SegFormer can be converted to an instance segmentation model by just discarding stuff queries.
In ~\cref{t2}, we report our instance segmentation results on COCO \texttt{test-dev} set. We achieve results comparable to the current state-of-the-art methods such as QueryInst~\cite{fang2021queryinst} and HTC~\cite{chen2019hybrid}, and 1.8 AP higher than K-Net~\cite{zhang2021k}. Using random crops during training boosts the AP by 1.3 percentage points.

\begin{table}
\renewcommand{\captionlabelfont}{\captionfontsize}
\renewcommand\tabcolsep{7pt}
\renewcommand\arraystretch{0.9}
\begin{center}
\resizebox{\linewidth}{!}{
\begin{tabular}{lccccc}
\toprule
Method & Backbone &${\rm AP}^{\rm seg}$ & ${\rm AP}^{\rm seg}_S$ & ${\rm AP}^{\rm seg}_M$ & ${\rm  AP}^{\rm seg}_L$   \\
\midrule
Mask R-CNN~\cite{he2017mask} & R50& 37.5 & 21.1 & 39.6 & 48.3\\ 
SOLOv2~\cite{wang2020solov2} & R50& 38.8 & 16.5 & 41.7 & 56.2 \\
K-Net~\cite{zhang2021k}  & R50 & 38.6 &19.1&  42.0 &\textbf{57.7}\\
SOLQ~\cite{dong2021solq} & R50 & 39.7 & 21.5 & 42.5 & 53.1 \\
HTC~\cite{chen2019hybrid} & R50& 39.7 & 22.6 & 42.2 & 50.6 \\
QueryInst~\cite{fang2021queryinst} & R50& 40.6 & \textbf{23.4} & 42.5 & 52.8 \\
\midrule
Ours (w/o crop) & R50& {40.4} & 21.1 & {43.8} & {54.7}\\
Ours (w/ crop) & R50 & \textbf{41.7} & 21.9 & \textbf{45.3} & {56.3} \\
\bottomrule
\end{tabular}

}
\end{center}
\vspace{-0.20in}  
\caption{\captionfontsize Instance segmentation on COCO \texttt{test-dev} set.} 
\label{t2}
\vspace{-0.1in} 
\end{table}

\vspace{-1mm}
\subsection{Ablation Studies}
\vspace{-1mm}
First, we show the effect of each module in \cref{stage}. Compared to baseline DETR, our model achieves better performance, faster inference speed and significantly reduces the training epochs. We use Panoptic SegFormer (R50) to perform ablation experiments by default.
\begin{table}[t]
\renewcommand{\captionlabelfont}{\captionfontsize}
\renewcommand\arraystretch{0.9}
\renewcommand\tabcolsep{4pt}
\begin{center}
\resizebox{\linewidth}{!}{
\begin{tabular}[b]{l c c c c c}

	\toprule
	& Epochs& PQ & \#Params & FLOPs & FPS \\
	\midrule
	baseline (DETR~\cite{carion2020end}) & 325 &43.4& 42.9M&247.5G & 4.9\\
	+ mask-wise merging & 325 &44.7 &42.9M &247.5G &6.1\\
	++ ms deformable attention~\cite{zhu2020deformable} & 50 &47.3 &44.9M &618.7G & 2.7 \\
	+++ mask decoder & 24 & 48.5 & 51.0M  &214.8G &  7.8\\
    ++++ query decoupling & 24 & 49.6& 51.0M& 214.2G& 7.8\\
    \bottomrule
\end{tabular}}
\end{center}
\vspace{-0.20in}  
\caption{\captionfontsize We increase the panoptic segmentation performance of DETR~\cite{carion2020end} (R50\cite{he2016deep}) from 43.4\% PQ to 49.6\% PQ with fewer training epochs, less computation cost, and faster inference speed.}
\label{stage}
\vspace{-0.2in}  
\end{table}

\setlength{\columnsep}{9pt}
\begin{wraptable}{r}{0.5\linewidth}
\renewcommand\tabcolsep{5.0pt} 
\vspace{-0.3in}
\scriptsize
\renewcommand\arraystretch{0.9}
\begin{center}
\scriptsize
	\resizebox{\linewidth}{!}{
	\scriptsize
        \begin{tabular}{l c c c}
\toprule
  \#Layer  & PQ & PQ$^{\rm th}$ & PQ$^{\rm st}$ \\
\midrule
0 & 47.0 &  50.0& 42.5\\
1 & 47.7 &  51.1& 42.5\\
2 & 48.1 & 51.8 & 42.5 \\
6* (box-free) & 49.2 & 53.5 & 42.6\\
6 & 49.6 & 54.4 & 42.4 \\
\bottomrule
\end{tabular}
    }
\end{center}
\vspace{-0.25in}
\caption{\scriptsize
Ablate location decoder.
}
 \label{det_apx}
\vspace{-0.2in}
\end{wraptable} 
\textbf{Effect of Location Decoder.}
Location decoder assists queries to capture the location information of things. 
\cref{det_apx} shows the results with varying the number of layers in the location decoder. With fewer location decoder layers, our model performs worse on things, and it demonstrates that learning location clues through the location decoder is beneficial to the model to handle things better. * notes we predict mass centers rather than bounding boxes in our location decoder, and this box-free model achieves comparable results (49.2\% PQ \vs. 49.6\% PQ).

\begin{figure}[t]
\renewcommand{\captionlabelfont}{\captionfontsize}
\begin{center}
   \includegraphics[width=\linewidth]{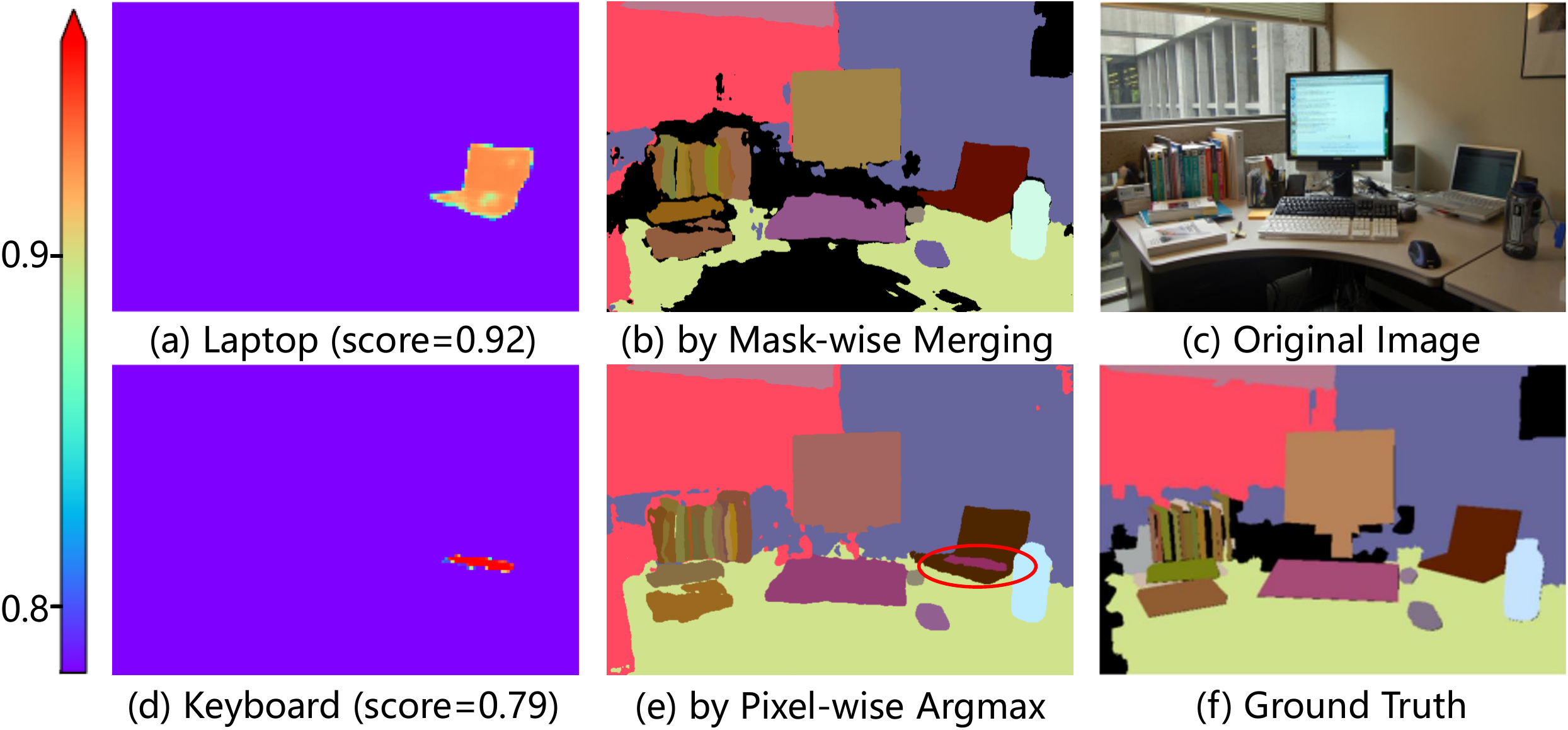}
\end{center} 
\vspace{-0.20in}  
   \caption{ \captionfontsize
   While using pixel-wise argmax, the keyboard is covered on the laptop (noted by the red circle in (e). However, the laptop has a higher classification probability than the keyboard. The pixel-wise argmax strategy fails to use this important clue.
   Masks logits were generated through DETR-R50~\cite{carion2020end}.
    }
\label{fig:laptop}
\vspace{-0.10in} 
\end{figure}

\textbf{Mask-wise Merging.}
As shows in ~\cref{post}, we compare our mask-wise merging strategy against pixel-wise argmax strategy on various models. We use both Mask PQ and Boundary PQ~\cite{cheng2021boundary} to make our conclusions more credible. Models with mask-wise merging strategy always performs better. DETR with mask-wise merging outperforms origin DETR by 1.3\% PQ~\cite{carion2020end}. In addition, our mask-wise merging is 20\% less time-consuming than DETR's pixel-wise argmax since DETR uses more tricks in its code, such as merging stuff with the same category and iteratively removing masks with small areas. \cref{fig:laptop} shows one typical fail case of using pixel-wise argmax.

\begin{table}[t]
\renewcommand{\captionlabelfont}{\captionfontsize}
\renewcommand\arraystretch{0.9}
\begin{center}
\resizebox{\linewidth}{!}{
\begin{tabular}{p{3.1cm}|p{0.75cm}<{\centering} p{0.75cm}<{\centering}p{0.75cm}<{\centering} |p{0.75cm}<{\centering} p{0.75cm}<{\centering}p{0.75cm}<{\centering}   }
 \toprule
 \multirow{2}{*}{Method}& \multicolumn{3}{c|}{Mask PQ} & \multicolumn{3}{c}{Boundary PQ\cite{cheng2021boundary}} \\

 &  PQ & SQ &  RQ & PQ & SQ & RQ   \\
\midrule
DETR (p) & 43.4 & 79.3 & 53.8 & 32.8 & 71.0 & 45.2 \\
DETR (m) & 44.7 & 80.2 & 54.7 &  33.7 & 71.1 & 46.5 \\

\midrule
D-DETR-MS (p) & 46.3 & 80.0 & 56.5 & 37.1 & 72.1 & 50.2 \\
D-DETR-MS (m) & 47.3 & 81.1 & 56.8 &  38.0 & 72.3 & 51.0 \\
\midrule
MaskFormer (p)& 45.6 & 80.2 & 55.8 & - & - &-  \\
MaskFormer (p*)  & 46.5 & 80.4 & 56.8 & 36.8 & 72.5 &49.8  \\
MaskFormer (m) & 46.8 & 80.4 & 57.2 & 37.6 & 72.6 &51.1 \\
\midrule
\small{Panoptic SegFormer (p)} & 48.4 & 80.7 & 58.9 & 39.3 & 73.0 &52.9  \\
\small{Panoptic SegFormer (m)}& 49.6 & 81.6 & 59.9 & 40.4 & 73.4 &54.2
\\
\bottomrule

\end{tabular}}
\end{center}
\vspace{-0.20in}  
\caption{\captionfontsize {Effect of mask-wise merging strategy.} The table shows the results of models with different post-processing methods, and the backbone is ResNet-50. ``(p)" refers to using pixel-wise argmax as the post-processing method.
``(p*)" considers both class probability and mask prediction probability in its pixel-wise argmax strategy~\cite{cheng2021per}. Models with ``(m)" that employ mask-wise merging always perform better in both Mask PQ and Boundary PQ~\cite{cheng2021boundary} than pixel-wise argmax method. 
}
\label{post}
\vspace{-0.10in}  
\end{table}

\begin{table}[t]
\renewcommand{\captionlabelfont}{\captionfontsize}
\renewcommand\arraystretch{0.9}
\begin{center}
\renewcommand\tabcolsep{6pt}

\resizebox{\linewidth}{!}{
\begin{tabular}{l c c c c c  }
\toprule
Method & {PQ} & {PQ}$^{\rm th}$&{PQ}$^{\rm st}$ & ${\rm AP}^{\rm box}$  &${\rm AP}^{\rm seg}$ \\
\midrule
DETR~\cite{carion2020end} & 43.4 & 48.2& 36.3 & 38.8 &31.1 \\
D-DETR-MS~\cite{zhu2020deformable} &  47.3& 52.6 & 39.0 & 45.3 & 37.6\\
Panoptic FCN~\cite{li2021fully} & 43.6 &49.3&35.0 & 36.6 &34.5 \\
Ours (Joint Matching) & 48.5&\textbf{54.5}&39.5 & 44.1 & 37.7\\
Ours (Query Decoupling)& \textbf{49.6} &54.4&\textbf{42.4}&\textbf{45.6} & \textbf{39.5}\\
\bottomrule

\end{tabular}
}
\end{center}
\vspace{-0.2in}
\caption{\captionfontsize Effect of query decoupling strategy. PQ and AP scores of various panoptic segmentation models on COCO \texttt{val2017}.
\vspace{-0.25in}
}
\label{inst_apx}
\end{table}
\textbf{Mask Decoder.} Our proposed mask decoder converges faster since the ground truth masks guide the attention module to focus on meaningful regions. \cref{fig:sche} shows the convergence curves of several models. We only supervise the last layer of the mask decoder while not employing deep supervision.
We can observe that our method achieves 49.6\% PQ with training for 24 epochs, and longer training has little effect. However, D-DETR-MS needs at least 50 epochs to achieve better performance. Deep supervision is vital for our mask decoder to perform better and converge faster. \cref{ds} shows the attention maps of different layers in the mask decoder, and the attention module focuses on the target car in the previous layer when using deep supervision. The attention maps are very similar to the final predicted masks, since masks are generated by attention maps with a  lightweight FC head.

\begin{figure}[t]
\renewcommand{\captionlabelfont}{\captionfontsize}
\begin{center}
   \includegraphics[width=0.9\linewidth]{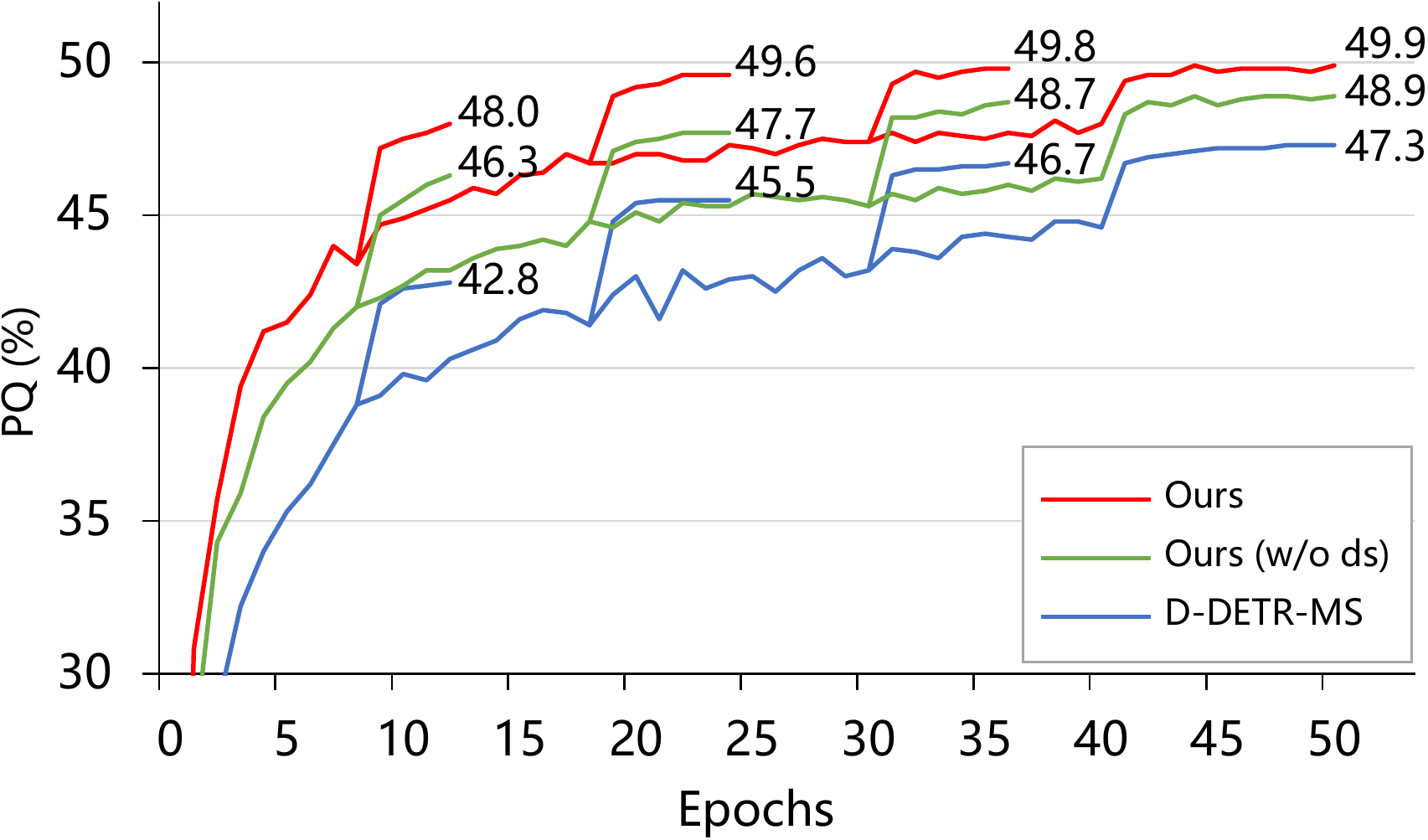}
\end{center}
\vspace{-0.20in}  
   \caption{\captionfontsize
   { Convergence curves of Panoptic SegFormer and D-DETR-MS.}
   We train models with different training schedules.
   ``w/o ds" refers that we do not employ deep supervision in the mask decoder.
    The learning rate is reduced where the curves leap.
    }
\label{fig:sche}
\vspace{-0.10in}  
\end{figure}
\renewcommand\tabcolsep{3.0pt}
\begin{figure}[t]
\renewcommand{\captionlabelfont}{\captionfontsize}
\centering
\resizebox{0.95\linewidth}{!}{
\begin{tabular}{c c  c  c  c }
    & Layer-1 & Layer-3 & Layer-6& Result\\
     \rotatebox{90}{\rule{5pt}{0pt} w/ ds}
    &
    \multicolumn{1}{c}{
    \begin{minipage}[b]{0.24\linewidth}
		\centering
	        {\includegraphics[width=\linewidth]{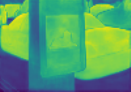}}
	        \end{minipage}
	        }
    & 
    \multicolumn{1}{c}{
    \begin{minipage}[b]{0.24\linewidth}
		\centering
		{\includegraphics[width=\linewidth]{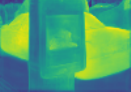}}
	\end{minipage}}
    & 
     \multicolumn{1}{c}{
    \begin{minipage}[b]{0.24\linewidth} 
		\centering
		{\includegraphics[width=\linewidth]{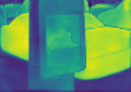}}
	\end{minipage}}
	&
	 \multicolumn{1}{c}{
	\begin{minipage}[b]{0.24\linewidth} 
		\centering
		{\includegraphics[width=\linewidth]{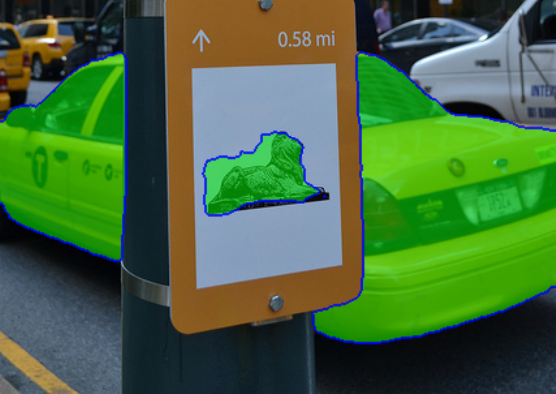}}
	\end{minipage}}
	\\
	\rotatebox{90}{\rule{5pt}{0pt}w/o ds}&
	\begin{minipage}[b]{0.24\linewidth}
		\centering
		{\includegraphics[width=\linewidth]{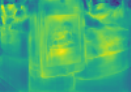}}
	\end{minipage}
	&
	\begin{minipage}[b]{0.24\linewidth}
		\centering
		{\includegraphics[width=\linewidth]{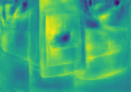}}
	\end{minipage}
	&
		\begin{minipage}[b]{0.24\linewidth}
		\centering
		{\includegraphics[width=\linewidth]{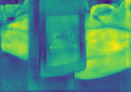}}
	\end{minipage}
	&
		\begin{minipage}[b]{0.24\linewidth}
		\centering
		{\includegraphics[width=\linewidth]{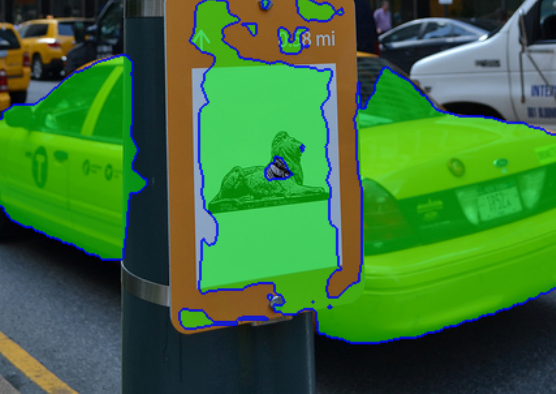}}
	\end{minipage}
	\\
  \end{tabular}
}
\vspace{-0.10in} 
\caption{ \captionfontsize Attention maps of different layers in the mask decoder. ``ds" refers to deep supervision.}
\label{ds}
\vspace{-0.20in} 
\end{figure}

\setcounter{table}{10}
\begin{table*}[h]
\renewcommand{\captionlabelfont}{\captionfontsize}
\renewcommand\arraystretch{1}
\begin{center}
\resizebox{\linewidth}{!}{\begin{tabular}{l|c c|cccc|cccc|cccc|cccc}
\toprule
\multirow{2}{*}{Method} & \multirow{2}{*}{Clean} &\multirow{2}{*}{Mean} &\multicolumn{4}{c|}{Blur} & \multicolumn{4}{c|}{Noise} & \multicolumn{4}{c|}{Digital} & \multicolumn{4}{c}{Weather} \\
\cline{4-19}
&  & & Motion & Defoc & Glass & \multicolumn{1}{c|}{Gauss} & Gauss & Impul & Shot & \multicolumn{1}{c|}{Speck} & Bright & Contr & Satur & \multicolumn{1}{l|}{JPEG} & \multicolumn{1}{l}{Snow} & \multicolumn{1}{l}{Spatt} & \multicolumn{1}{l}{Fog} & \multicolumn{1}{l}{Frost} \\
\midrule
\multicolumn{1}{l|}{Panoptic FCN (R50)} & \multicolumn{1}{c}{43.8}& \multicolumn{1}{c|}{26.8} & 22.5 & 23.7 & 14.1 & \multicolumn{1}{c|}{25.0}& 28.2 & 20.0 & 28.3 & \multicolumn{1}{c|}{31.9}& 39.4 & 24.3 & 38.0 & \multicolumn{1}{c|}{22.9} & 20.0 & 29.6 & 35.3 & 25.3 \\
\multicolumn{1}{l|}{MaskFormer (R50)} &
\multicolumn{1}{c}{47.0}& \multicolumn{1}{c|}{29.5} & 24.9 & 28.1 & 16.4 & \multicolumn{1}{c|}{29.5} & 31.2 & 24.7 & 30.9 & \multicolumn{1}{c|}{34.8} & 42.5 & 27.5 & 41.2 & \multicolumn{1}{c|}{22.0} & 20.4 & 31.0 & 38.5 & 27.7 \\
\multicolumn{1}{l|}{D-DETR (R50)} &
\multicolumn{1}{c}{47.6}& \multicolumn{1}{c|}{30.3} & 25.6 & 28.7 & 16.8 & \multicolumn{1}{c|}{29.7} & 32.5 & 24.9 & 31.4 & \multicolumn{1}{c|}{35.9} & 43.1 & 28.6 & 41.3 & \multicolumn{1}{c|}{24.5} & 21.7 & 31.7 & 39.7 & 28.7 \\
\multicolumn{1}{l|}{Ours (R50)} &
\multicolumn{1}{c}{50.0}&\multicolumn{1}{c|}{32.9} & 26.9 & 30.2 & 17.5 & \multicolumn{1}{c|}{31.6} & 35.5 & 27.9 & 35.4 & \multicolumn{1}{c|}{38.6} & 45.7 & 31.3 & 43.9 & \multicolumn{1}{c|}{29.0} & 24.3 & 35.0 & 41.9 & 32.3 \\
\midrule
\multicolumn{1}{l|}{MaskFormer (Swin-L)} & \multicolumn{1}{c}{52.9}&\multicolumn{1}{c|}{41.7} & 37.3 & 38.0 & 30.4 & \multicolumn{1}{c|}{39.3} & 42.3 & 42.5 & 42.8 & \multicolumn{1}{c|}{45.3} & 49.7 & 43.9 & 49.4 & \multicolumn{1}{c|}{39.7} & 35.2 & 45.2 & 48.8 & 37.9 \\
\multicolumn{1}{l|}{Ours (Swin-L)} & 
\multicolumn{1}{c}{\textbf{55.8}}&\multicolumn{1}{c|}{\textbf{47.2}} & 41.3 & 41.5 & 34.3 & \multicolumn{1}{c|}{42.7} & 48.6 & 49.5 & 48.8 & \multicolumn{1}{c|}{50.3} & 53.8 & 50.1 & 53.5 & \multicolumn{1}{c|}{46.9} & 44.8 & 51.5 & 53.3 & 44.3 \\
\midrule
\multicolumn{1}{l|}{Ours (PVTv2-B5)} & \multicolumn{1}{c}{55.6} & \multicolumn{1}{c|}{47.0}& 41.5&41.1&36.1&42.5&48.4&49.6&48.4&50.4&53.5&50.8&53.0&46.2&42.4&50.3&52.9&44.3\\
\bottomrule
\end{tabular}

}
\end{center}
\vspace{-0.2in}
\caption{\captionfontsize
    Panoptic segmentation results on COCO-C. To ease the workload of the experiment, we use a subset of 2000 images from the COCO \texttt{val2017}. The third column is the average results on 16 types of corruption data. 
}
\label{coco_c}
\vspace{-0.20in}
\end{table*}

\setcounter{table}{9}
\setlength{\columnsep}{9pt}
\renewcommand\tabcolsep{5.0pt} 
\begin{wraptable}{r}{0.5\linewidth}
\scriptsize
\vspace{-0.3in}
\begin{center}
\resizebox{\linewidth}{!}{
\begin{tabular}{l c c c c}
\toprule
 Layer & PQ & PQ$^{\rm th}$ & PQ$^{\rm st}$ & Fps\\
\midrule
1st  & 48.8 & 54.3 & 40.5 & 10.6 \\
2nd  & 49.5 & 54.5 & 42.0 & 9.8\\
3rd  & 49.6 & 54.5 & 42.3 & 9.3\\
Last & 49.6 & 54.4 & 42.4 & 7.8\\
\bottomrule
\end{tabular}}
\end{center}
\vspace{-0.25in}
\caption{\scriptsize
    Results of each layer in the mask decoder.
}
\label{light}
\vspace{-6mm}
\end{wraptable}

Since our mask decoder can generate masks from each layer, we evaluate the performance of each layer in the mask decoder, see~\cref{light}. During inference, using the first two layers of mask decoder will be on par with the whole mask decoder. It also inferences faster because the computational cost decreases. PQ$^{\rm th}$ is hardly affected by the number of layers, PQ$^{\rm st}$ performs a little poorly in the first layer. The reason is that the location decoder has made additional refinements to the thing queries.

\textbf{Effect of Query Decoupling Strategy.}
We compare our proposed query decoupling strategy with previous DETR's matching method (described here as ``joint matching")~\cite{carion2020end}, as shown in ~\cref{inst_apx}. 
Following DETR, joint matching uses a set of queries to target both things and stuff and feeds all queries to both location decoder and mask decoder. For our proposed query decoupling strategy, we use thing queries to detect things through bipartite matching and use location decoder to refine them. Stuff queries are assigned through class-fixed assign strategy. For a fair comparison, both the joint matching strategy and our query decoupling strategy employ 353 queries. 
We can observe that our proposed strategy highly boost PQ$^{\rm st}$. In addition, panoptic segmentation model can perform instance segmentation by utilizing its thing results only. However, previous panoptic segmentation methods always perform poorly on instance segmentation task even though the two tasks are closely related.
\cref{inst_apx} shows both panoptic segmentation and instance segmentation performance of various methods. Our query decoupling strategy can achieve sota performance on panoptic segmentation task while obtaining a competitive instance segmentation performance.
\renewcommand{\captionlabelfont}{\scriptsize}
\begin{wrapfigure}{r}{0.45\linewidth}
  \vspace{-0.3in}
  \renewcommand\arraystretch{0.9}
  \begin{center}
  \advance\leftskip+1mm
  \renewcommand{\captionlabelfont}{\scriptsize}
    \includegraphics[width=\linewidth]{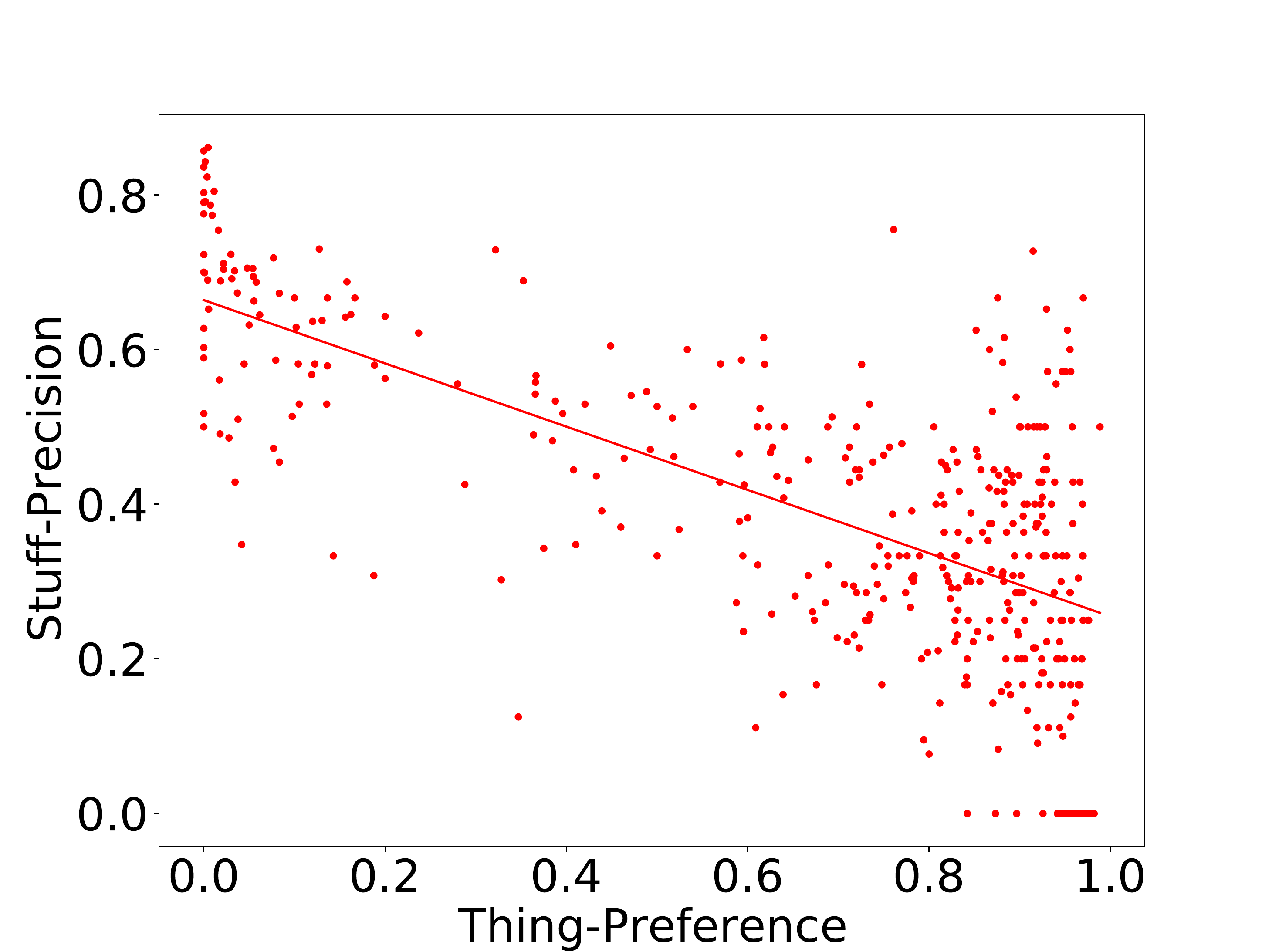}
    \vspace{-0.3in} 
    \caption{\scriptsize Things-Preference \vs. Stuff-Precision. }\label{thing-stuff}
  \end{center}
  \vspace{-0.3in}
\end{wrapfigure}
In short, query decoupling strategy achieves higher PQ$^{\rm st}$ and AP$^{\rm seg}$ compared to joint matching. We analyze the experimental results of joint matching and find that if one query prefers things more, the precision of stuff results detected by it will be lower, see \cref{thing-stuff}. Each point represents the Thing-Preference and Stuff-Precision corresponding to each query, and the specific definitions are in Appendix. The red line is the linear regression of these points. When using one query set to detect things and stuff together, it will cause interference within each query. Our query decoupling strategy prevents things and stuff from interfering within the same query.

\vspace{-2mm}
\subsection{Robustness to Natural Corruptions}
\vspace{-1mm}
Panoptic segmentation has promising applications in many fields, such as autonomous driving. Model robustness is one of the top concerns of autonomous driving. In this experiment, we evaluate the robustness of our model to disturbed data. We follow ~\cite{kamann2020benchmarking} and generate COCO-C, which extends the COCO validation set to include disturbed data generated by 16 algorithms from  blur, noise, digital and weather categories. We compare our model to Panoptic FCN~\cite{li2021fully}, D-DETR-MS and MaskFormer~\cite{cheng2021per}. The results are shown in~\cref{coco_c}. We calculated the mean results of disturbed data on COCO-C. Using the same backbone, our model always performs better than others. Previous literature~\cite{xie2021SegFormer,bhojanapalli2021understanding,naseer2021intriguing} found that transformer-based model has stronger robustness on image classification and semantic segmentation tasks. Our experimental results also show that the transformer-based backbone (Swin-L and PVTv2-B5) can bring better robustness to the model. However, for tasks requiring a more complex pipeline, such as panoptic segmentation, we argue that the design of the task head also plays an important role for the robustness of the model. For example, Panoptic SegFormer (Swin-L) has an average result of 47.2\% PQ on COCO-C, outperforming MaskFormer  (Swin-L) by 5.5\% PQ, higher than their gap (2.9\% PQ) on clean data. We posit it is due to our transformer-based mask decoder having stronger robustness than the convolution-based pixel decoder of MaskFormer.

\vspace{-2mm}
\section{Conclusion}
\vspace{-1mm}

\textbf{Limitation.} This work relies on deformable attention to process multi-scale features, and the speed is a little slow. Our model is still hard to handle features with a larger spatial shape and does not perform well for small targets.

\textbf{Discussion.} Recently, the segmentation field attempted to use a uniform pipeline to process various tasks, including semantic segmentation, instance segmentation, and panoptic segmentation. However, we think that complete unification is conceptually exciting but not necessarily a suitable strategy. 
Given the similarities and differences among the various segmentation tasks, ``seek common ground while reserving differences" is a more reasonable guiding ideology. With query decoupling strategy, we can handle things and stuff in the same paradigm since they are represented as queries. In addition, we can also design customized pipelines for things or stuff. Such a flexible strategy is more suitable for various segmentation tasks. 
At present, task-specific designs still bring better performance. We encourage the community to further explore the unified segmentation frameworks and expect that Panoptic SegFormer can inspire future works.

\section{Acknowledge}
This work is supported by the Natural Science Foundation of China under Grant 61672273 and Grant 61832008.
Ping Luo is supported by the General Research Fund of HK No.27208720 and 17212120. Wenhai Wang and Tong Lu are corresponding authors.

\appendix

{\noindent\Large\textbf{Appendix}}
\setcounter{page}{1}
\setcounter{figure}{0}
\setcounter{table}{0}
\counterwithin{figure}{section}
\counterwithin{table}{section}

\renewcommand{\captionlabelfont}{\normalsize}
\section{Implementation Details}\label{details}
\subsection{Panoptic SegFormer.}
Our settings mainly follow DETR~\cite{carion2020end} and Deformable DETR~\cite{zhu2020deformable} for simplicity. The hyper-parameters in deformable attention are the same as Deformable DETR~\cite{zhu2020deformable}.
We use Channel Mapper~\cite{mmdetection,zhu2020deformable} to map dimensions of the backbone's outputs to 256. 
The location decoder contains 6 deformable attention layers, and the mask decoder contains 6 vanilla cross-attention layers~\cite{vaswani2017attention}. The spatial positional encoding is the commonly used fixed absolute encoding that is the same as DETR. The window size of Swin-L~\cite{liu2021swin} we used is 7.
 Since we equally treat each query.
$\lambda_{\rm things}$ and $\lambda_{\rm stuff}$ are dynamically adjusted according to the relative proportion of things and stuff in each image, and their sum is 1. $\lambda_{\rm cls}$, $\lambda_{\rm seg}$, and $\lambda_{\rm det}$ in \cref{things_loss} are set to 2, 1, 1, respectively. 

During the training phase, the predicted masks that be assigned $\varnothing$ will have a weight of zero in computing $\mathcal{L}_{seg}$.
While using the mass center of instance to replace the bounding box, we only use L1 loss to supervise the mass center of predicted mask and mass center of ground truth.
We employ a threshold 0.5 to obtain binary masks from soft masks. Threshold ${\rm t}_{\rm cnf}$ and ${\rm t}_{\rm keep}$ are 0.25 (0.3) and 0.6, respectively. $\alpha$ and $\beta$ in \cref{confidence} are 1 and 2, respectively. All experiments are trained on one NVIDIA DGX node with 8 Tesla V100 GPUs.

By default, for COCO dataset~\cite{lin2014microsoft}, We train our models with 24 epochs, a batch size of 1 per GPU, a learning rate of $1.4
\!\times\! 10^{-4}$ (decayed at the 18th epoch by a factor of 0.1, learning rate multiplier of the backbone is 0.1). We use a multi-scale training strategy with the maximum image-side not exceeding 1333 and the minimum image size varying from 480 to 800, and random crop augmentations is applied during training. The number of thing queries $N_{\rm th}$ is set to 300. Stuff queries have tge equal number of stuff classes, and it is 53 in COCO. 

For the ADE20K dataset~\cite{zhou2017scene}, we train our model with 100 epochs (decayed at 80th epoch), image size varying from 512 to 2048. Since ADE20K contains 50 stuff, we use 50 stuff queries.  Other settings are the same to COCO dataset.

\textbf{FPS and FLOPs.} FPS in Tab.5 is measured on a V100
GPU with a batch size of 1. "DETR" and "DETR+mask wise merging" are from Detectron2~\cite{wu2019detectron2} and DETR's implementation. Others are from Mmdet~\cite{mmdetection} and our own implementation. Our framework is slightly more efficient than DETR. FLOPs of DETR are measured from Detectron2  on an average of 100 images.

\subsection{Deformable DETR for Panoptic Segmentation}
Following DETR for panoptic segmentation, we transplanted the panoptic head of DETR to Deformable DETR.
To ensure consistency, we only generate the attention maps with the spatial shape of 32s.
When using single scale deformable DETR, the process of generating attention maps is the same as DETR.  When using multi-scale deformable DETR, we only multiply queries and the features (from C5) to generate attention maps. Other settings of deformable DETR for object detection are kept unchanged. We apply iterative bounding box refinement as the default setting for Deformable DETR.  We use 300 queries and this brings huge computation costs, although this model achieves pretty good performance.
\section{Discussion}

We will deliver more ablation studies, more detailed analysis in this section.

\setlength{\columnsep}{4pt}
\begin{wraptable}{r}{0.5\linewidth}
	\scriptsize
	\vspace{-0.30in}
	\renewcommand\arraystretch{0.9}
	\begin{center}
    \resizebox{\linewidth}{!}{
    \begin{tabular}{lcccc}
	\toprule
	Method& Epoch& PQ & PQ$^{\rm th}$ & PQ$^{\rm st}$  \\
	\midrule
	DETR~\cite{carion2020end} & 500 & 43.4 & 48.2 &36.3\\
    D-DETR-SS & 50 & 40.6 & 44.0 & 35.4 \\
    D-DETR-MS & 50 & 46.3 & 51.9 & 37.9 \\
    \bottomrule
    \end{tabular}
    }
    \end{center}
      \vspace{-0.25in}
    \caption{\normalsize   ``D-", ``SS" and ``MS" refers to ``Deformable", single-scale and multi-scale.
    }
    \label{deformable}
    \vspace{-0.2in}
\end{wraptable}
\textbf{Effect of Deformation Attention.}
To ablate the effect of deformable attention, we extend Deformable DETR on panoptic segmentation with the panoptic head of DETR. For more implementation details, please refer to the \cref{details}.
As shown in ~\cref{deformable}, multi-scale deformable attention improves  2.9\% PQ compared to DETR. Multi-scale attention outperforms single-scale attention by 5.7\% PQ, highlighting the important role of multi-scale features for segmentation task.

\subsection{Post-processing Method}\label{ppm}
\begin{figure}[t]
\begin{center}
   \includegraphics[width=\linewidth]{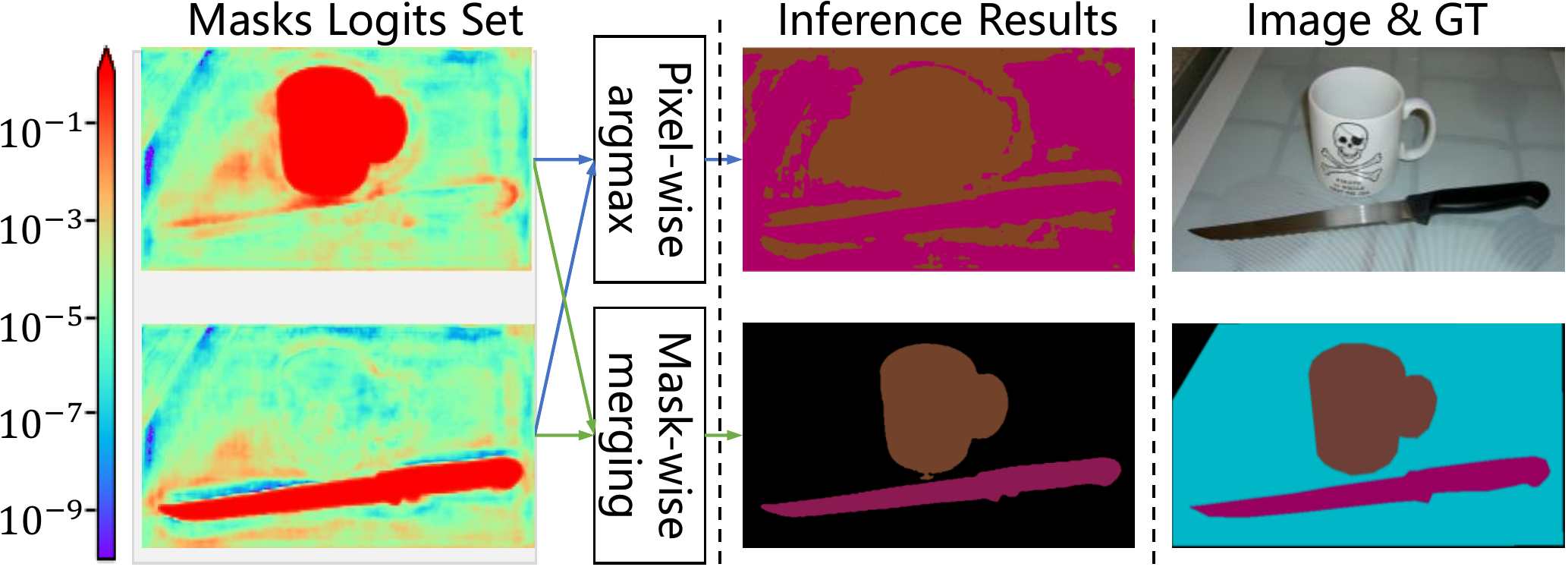}
\end{center}
\vspace{-0.2in}
   \caption{ \normalsize  
   \textbf{Pixel-wise Argmax \vs Mask-wise Merging.} We use DETR-R50 to compare the results generated through pixel-wise armgax and mask-wise merging. Firstly, DETR-R50 detects a cup and a knife from the image. When using pixel-wise argmax, other pixels ( dining table) are incorrectly filled with "cup" or "knife". It mistakenly believes that the largest mask logit is the correct result, regardless of its value. However, our mask-wise merging strategy generates the correct results since we binarize each mask. 
    }
    \vspace{-0.1in}
\label{fig:2592}
\end{figure}

\begin{figure}[t]
\centering
\includegraphics[width=0.9\linewidth]{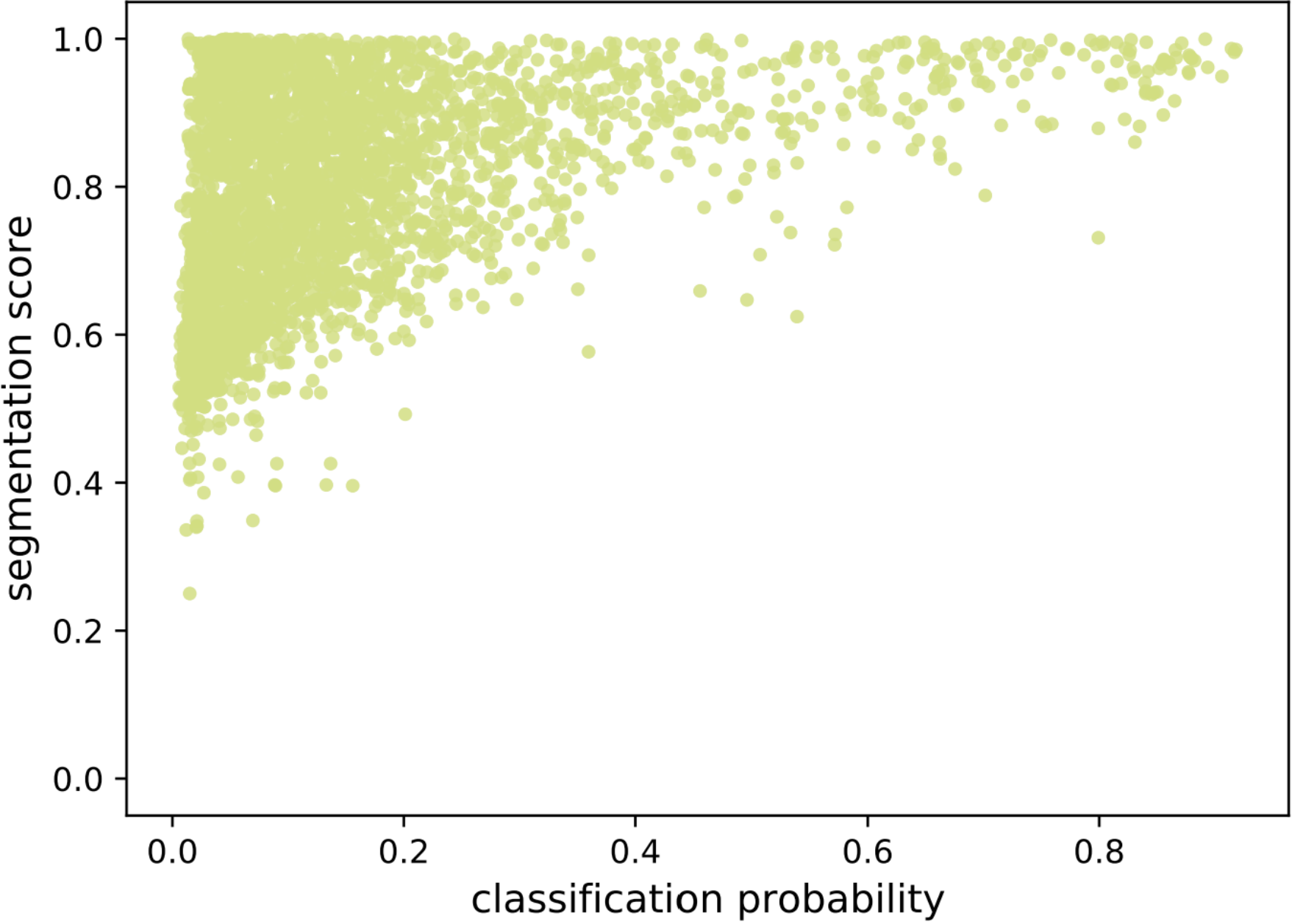}
\vspace{-0.1in}
\caption{ \normalsize  
The joint distribution for classification probability and segmentation score. We can observe that segmentation scores can be high while the masks have low classification probability.
}

\vspace{-0.1in}
\label{mask_cls}
\end{figure}
\begin{figure*}[t]
\centering
\includegraphics[width=0.7\linewidth]{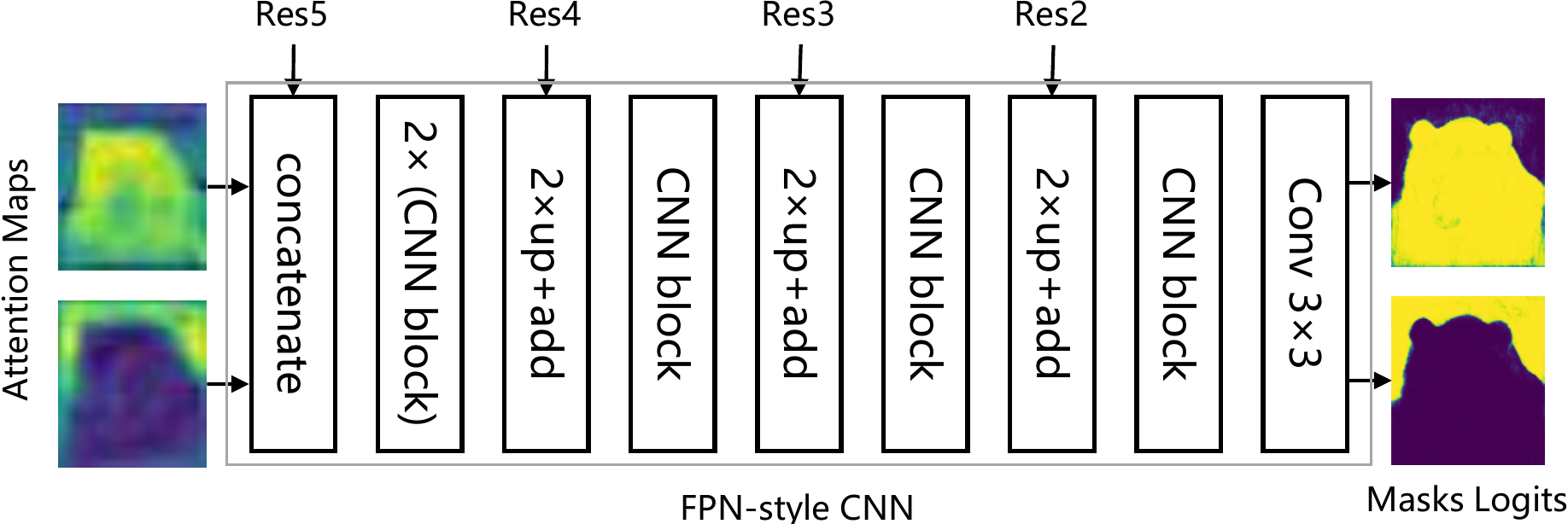}
\vspace{-0.1in}
\caption{  \normalsize   Illustration of DETR's panoptic head. CNN block consists of $3\!\times\!3$ convolution, GN, and ReLU. }
\label{panoptic_head}
\vspace{-0.1in}
\end{figure*}
\textbf{Defects of Pixel-wise Argmax.}\label{armgax} Pixel-wise argmax only considers the mask logits of each pixel. It has multiple issues that may lead to incorrect results. First of all, the pixel value generated from argmax may be extremely small, as shown in \cref{fig:2592}, which will generate plenty of false-positive results. The second issue is that the pixel with max mask logit may be the suboptimal result, as shown in \cref{fig:laptop} of the paper. 
This kind of error frequently appears in the segmentation maps generated by pixle-wise argmax. MaskFormer~\cite{cheng2021per} alleviates this problem by multiplying the classification probability by the masks logits. But this kind of error will still exist.

\begin{figure}[t]
\begin{center}
   \includegraphics[width=0.8\linewidth]{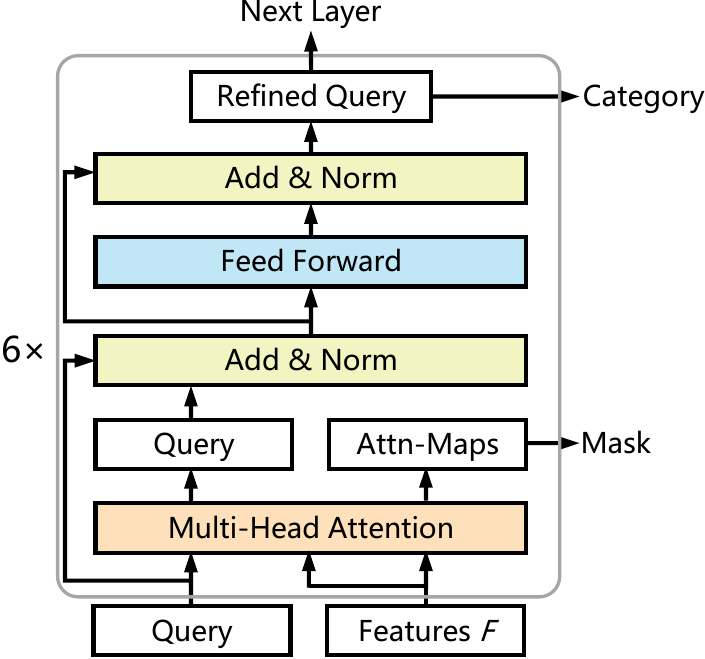}
\end{center}
\vspace{-0.2in}
   \caption{ \normalsize 
   \textbf{Architecture of mask decoder.} Attn-Maps notes attention maps.
   }
  \label{fig:mask_decoder_apx}
  
\end{figure} 

\begin{figure}[t]
\begin{center}
   \includegraphics[width=0.8\linewidth]{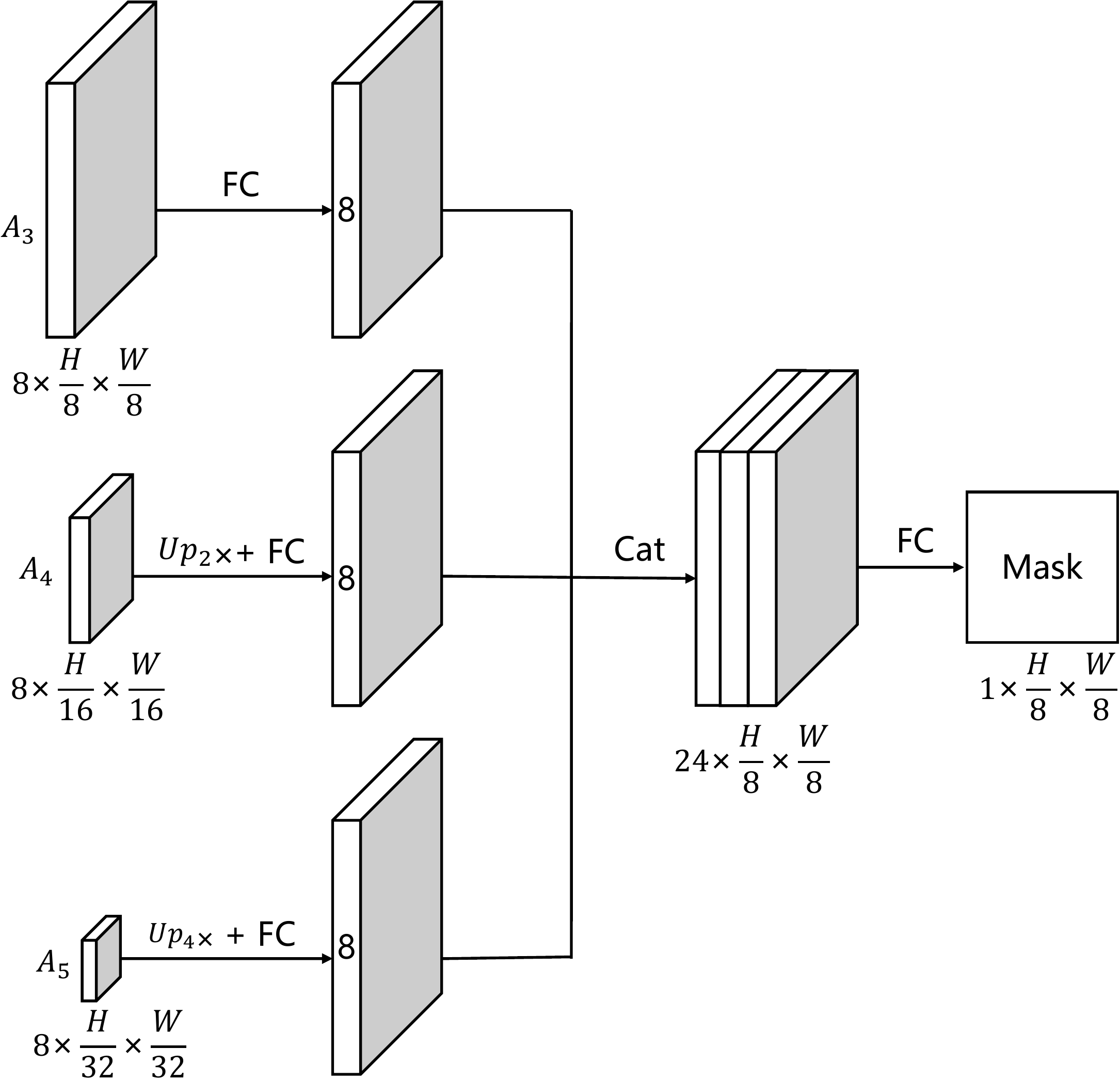}
\end{center}
\vspace{-0.2in}
   \caption{ \normalsize 
   Illustration of the module that generates mask from multi-scale multi-head attention maps. Up$_{2x}$ means upsampling by two times. FC notes fully connected layer. Cat notes concatenate.  While using 8 heads in the attention module, this module only contains 200+ parameters.
   }
   \vspace{-0.1in}
\label{fig:mixer_apx}
\end{figure}
\begin{table}[t]
\begin{center}
\renewcommand\tabcolsep{5pt} 
\begin{tabular}{l|c c c}

Post-Processing Method  & PQ & PQ$^{\rm th}$ &  PQ$^{\rm st}$ \\
\hline
Pixel-wise Argmax &  48.4 & 53.2 &41.3 \\
Heuristic Procedure\cite{kirillov2019panoptic} & 48.4& 54.3& 39.4\\
Mask-wise Mering  & 49.6 & 54.4& 42.4\\

\end{tabular}
\end{center}
\vspace{-0.2in}
\caption{ \normalsize
    The results of Panoptic SegFormer (R50) with different post-processing methods. Because the heuristic procedure always prefers things, it has the  lowest PQ$^{\rm st}$
}
\label{post2_apx}
\vspace{-0.1in}
\end{table}

\begin{table}[t]
\begin{center}
\renewcommand\tabcolsep{10pt} 
\begin{tabular}{c c|c c c}
  $\alpha$ & $\beta$& PQ & PQ$^{\rm th}$ & PQ$^{\rm st}$ \\
\hline
1 & 0& 48.7 & 53.5 & 41.3 \\
0 & 1 & 44.4 & 52.1 & 32.7 \\
1 & 1 & 49.3 & 54.1 & 42.1 \\
1 & 2 & 49.6 & 54.4 & 42.4\\
1 & 3 & 49.7 & 54.5 & 42.2\\
\end{tabular}
\end{center}
\vspace{-0.2in}
\caption{ \normalsize
    The weights of the  classification score and segmentation score determine the priority of masks. We can observe that employing both of them will perform better. According to the results on multiple models, we choose $\alpha\!=\!1$, $\beta\!=\!2$ as our default setting.
}
\label{post_apx}
\vspace{-0.1in}
\end{table}

\textbf{Heuristic Procedure.}
The heuristic procedure~\cite{kirillov2019panoptic} was the first proposed post-processing method of panoptic segmentation. It uses different strategies to handle things and stuff separately. Pixel-wise argmax was still used in its stuff workflow. One apparent  defect of this method is that it solves the overlap problem of stuff and things by always preferring things. This is an unfair way of dealing with stuff. \cref{post2_apx} shows that PQ$^{\rm st}$ of using heuristic procedure is lower than other methods because all stuff are treated unfairly.

\textbf{Masks-wise Merging.} 
Post-processing of panoptic segmentation aims to solve the overlap problem between masks. Although pixel-wise argmax uses an intuitive method to solve the overlap problem, it has defects mentioned above. We solve the overlap problem by giving different masks different priorities.
Mask-wise merging guarantees that high-quality masks have higher priority by sorting the masks with confidence scores. This strategy ensures that low-quality instances will not cover high-quality instances. In order to be able to effectively distinguish the quality of the masks, we consider both classification probability and segmentation score as the confidence score of each mask. The segmentation score ${\rm average (\mathbbm{1}_{\{m_i[h,w]>0.5\}}m_i[h,w])}$ represents the confidence of the overall segmentation quality of the mask. \cref{post_apx} shows the results of varying $\alpha$ and $\beta$ in Eq.6. Applying both classification probability and segmentation scores always have better performance. \cref{mask_cls} shows the relationship of classification probability and segmentation score. While one mask has a low classification probability ($[0,0.4]$), it may have  a large segmentation score. Large segmentation score means it has many pixels with high logits and this may generate false-positive results through pixel-wise argmax since its classification probability is pretty low.

Our mask-wise merging needs two thresholds to filter out undesirable results.
\cref{thr} shows that our algorithm is not very dependent on the choice of threshold t$_{\rm cnf}$ and t$_{\rm keep}$.
\cref{thr}
\begin{table}[t]
\begin{center}
\renewcommand\tabcolsep{6pt} 
\begin{tabular}{c| c c c c c}
\diagbox{t$_{\rm cnf}$}{t$_{\rm keep}$}& 0.9 &  0.8& 0.7 & 0.6 & 0.5\\
\hline
0.20	&48.9&  \textbf{49.5}&	\textbf{49.6}&	\textbf{49.6}&	49.4\\
0.25&	48.9&	\textbf{49.6}&	\textbf{49.7}&	\textbf{49.7}&	\textbf{49.5}\\
0.30	&48.8&	\textbf{49.5}&	\textbf{49.6}&	\textbf{49.6}&	\textbf{49.5}\\
0.35	&48.3&	49.1&	49.2&	49.2&	49.1\\
0.40&	47.4&	48.1&	48.2&	48.2&	48.1

\end{tabular}
\end{center}
\vspace{-0.2in}
\caption{\normalsize
   We use two thresholds t$_{\rm cnf}$ and t$_{\rm keep}$ in our mask-wise merging. We evaluate the results by combining different thresholds with Panoptic SegFormer (R50) to verify whether our algorithm is sensitive to these thresholds. Results higher than 49.5 are displayed in bold.
}
\vspace{-0.1in}
\label{thr}
\end{table}

\begin{figure*}[t]
\begin{center}
   \includegraphics[width=0.95\linewidth]{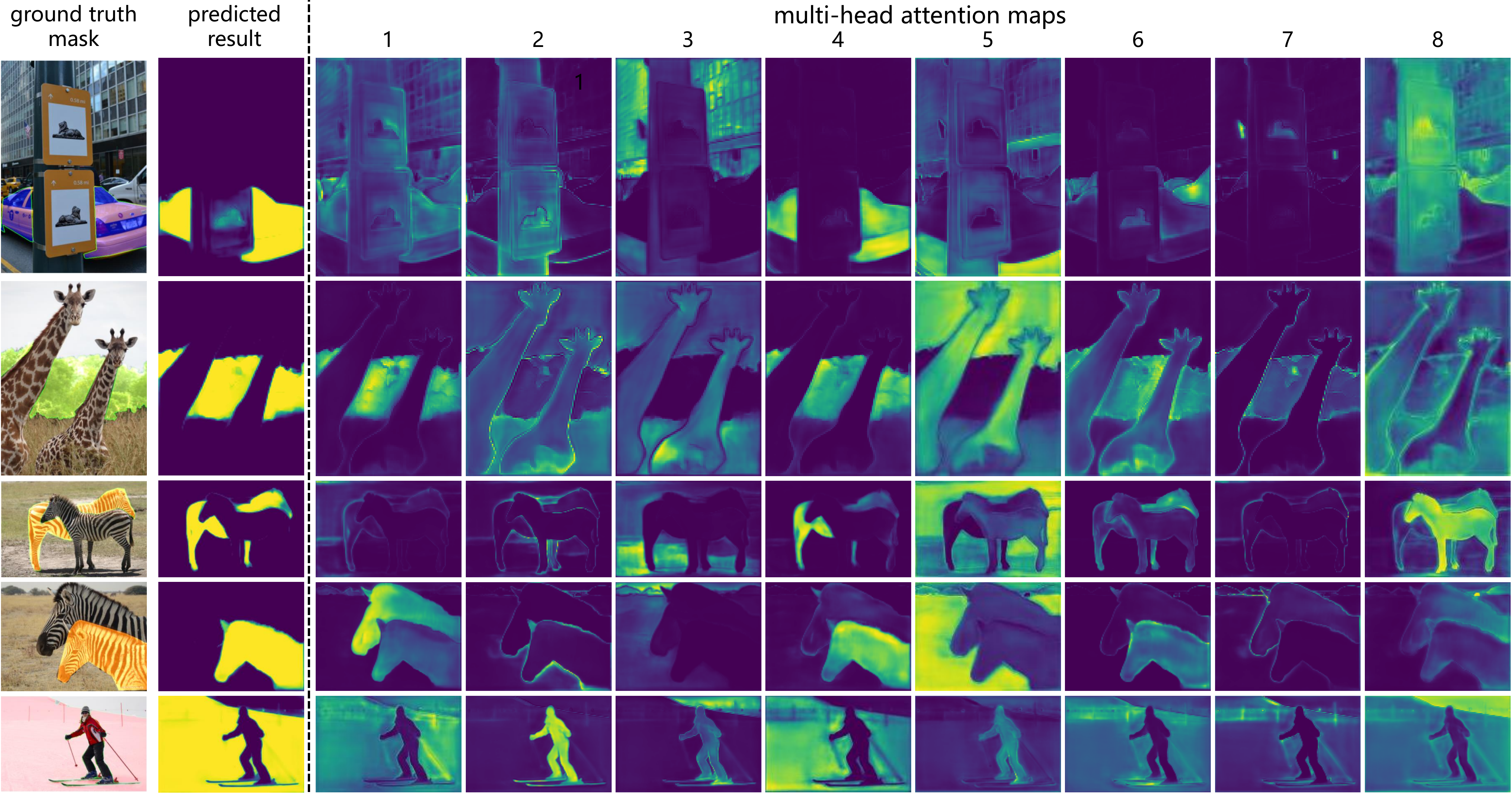}
\end{center}
\vspace{-0.2in}
   \caption{ \normalsize  
 \textbf{Visualization of multi-head attention maps and corresponding outputs from mask decoder.} Different heads have different preferences. Head 4 and Head 1 pay attention to foreground regions, and Head 8 prefers regions that occlude foreground. Head 5 always pays attention to the background that is around the foreground. Through the collaboration of these heads, Panoptic SegFormer can predict accurate masks. The 3rd row shows an impressive result of a horse that is highly obscured by the other horse.
   }
\label{fig:mask}
\end{figure*}
\begin{figure}[t]
\begin{center}
   \includegraphics[width=0.9\linewidth]{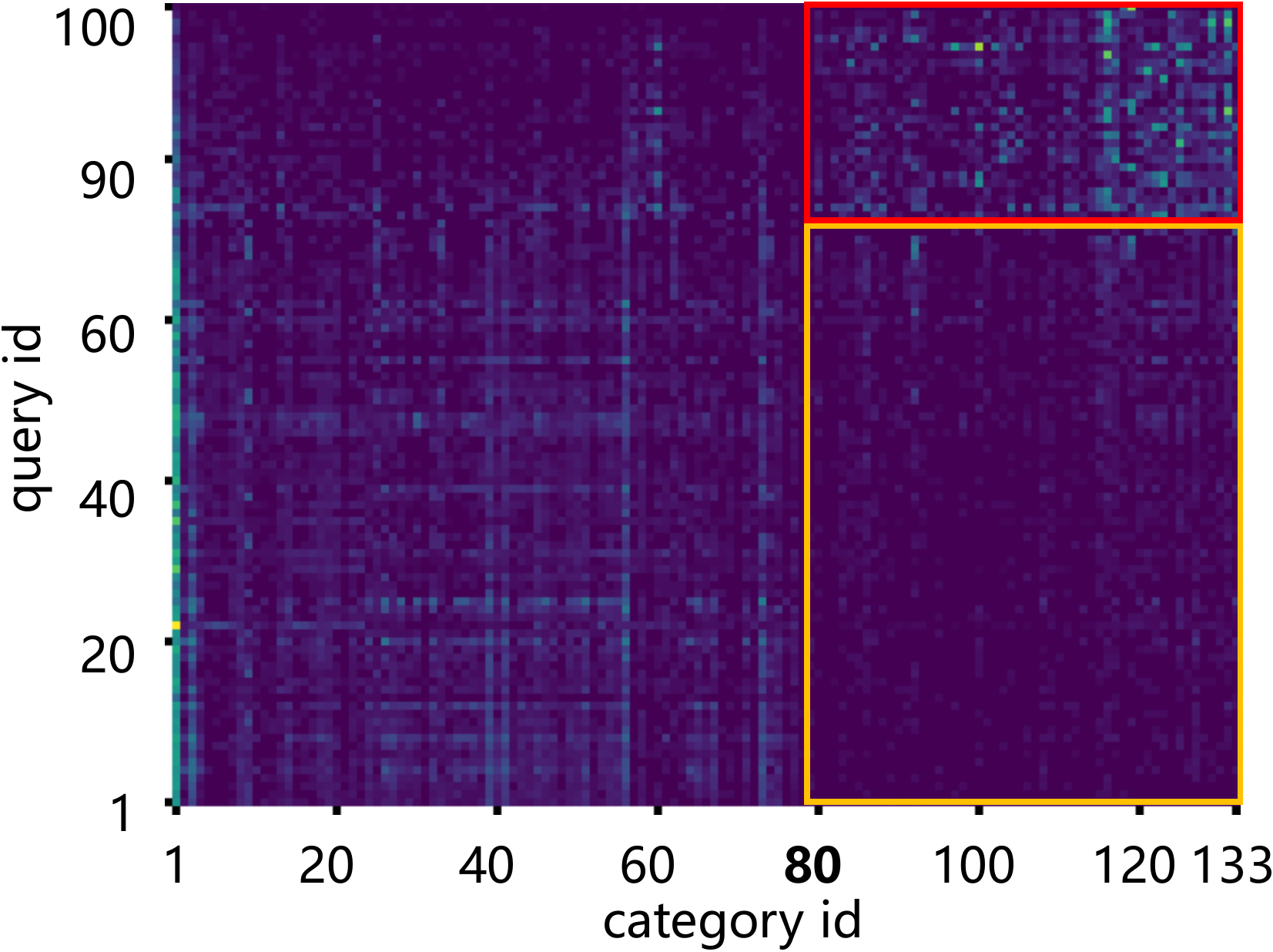}
\end{center}
\vspace{-0.2in}
   \caption{ \normalsize 
   \textbf{The Joint Distribution for Queries and Category in DETR.}
    We can observe that queries prefer either things or stuff. Although a few queries predict most of the stuff results ( within the red box), other queries still generate a considerable proportion of stuff results (within the yellow box). Our experimental results demonstrate that the results in the yellow box are usually of low quality.
    We sort the query ids for better visualization. Other literature~\cite{wang2021max} reports similar phenomenon.
    }
\label{fig:detr_query}
\end{figure}
\begin{table}
\begin{center}
\renewcommand\tabcolsep{10pt} 
\begin{tabular}{l c c c}
\toprule
\#Head & PQ & PQ$^{\rm th}$ &  PQ$^{\rm st}$ \\
\midrule
1 &  49.2 & 54.0 &42.0 \\
8 & 49.6 & 54.4& 42.4\\
\bottomrule

\end{tabular}

\vspace{-0.2in}
\end{center}
\caption{
    We varied  the number of heads in our mask decoder. More heads can bring slight  performance improvements. 
}

\label{head}
\end{table}

\begin{table*}
\begin{center}
\renewcommand\tabcolsep{10pt} 
\begin{tabular}{c |c | c c c |c c c  }
 \multirow{2}{*}{P$_{t}$} & \multirow{2}{*}{ \#Query} &\multicolumn{3}{c|}{Stuff} & \multicolumn{3}{c}{Things} \\
&  & TP &  TP+FP & Precision  & TP & TP+FP & Precision   \\
\hline
$\left[0.0,~0.1\right)$ & 44 & 7318 & 10060 & 0.73 & 136 & 222 & 0.61\\
$\left[0.1,~0.2\right)$ & 17 & 839 & 1308 & 0.64 & 140 & 198 & 0.71\\
$\left[0.2,~0.3\right)$ & 4 & 121 & 212 & 0.57 & 53 & 69 & 0.77\\
$\left[0.3,~0.4\right)$ & 11 & 339 & 646 & 0.52 & 252 & 368 & 0.68\\
$\left[0.4,~0.5\right)$ & 10 & 211 & 446 & 0.47 & 212 & 365 & 0.58\\
$\left[0.5,~0.6\right)$ & 15 & 327 & 684 & 0.48 & 477 & 903 & 0.53\\
$\left[0.6,~0.7\right)$ & 24 & 339 & 810 & 0.42 & 1001 & 1465 & 0.68\\
$\left[0.7,~0.8\right)$ & 40 & 400 & 1094 & 0.37 & 2019 & 3255 & 0.62\\
$\left[0.8,~0.9\right)$ & 83 & 539 & 1586 & 0.34 & 6325 & 9687 & 0.65\\
$\left[0.9,~1.0\right]$ & 105 & 309 & 1029 & 0.30& 11252 & 16724 & 0.67\\
\hline
Total & 353 & 10742 & 17875 &0.60 & 21867 &33256& 0.66 \\
\end{tabular}
\end{center}
\vspace{-0.2in}
\caption{ \normalsize
 We divide 353 queries into ten groups according to their P$_t$. For each group, we calculate their precision on stuff and things. Queries with higher P$_{t}$ have very low precision when they predict stuff.  
 This demonstrates that things may interfere with the prediction of stuff and using queries to target both things and stuff is suboptimal.
}
\label{query_apx}
\end{table*}

\begin{figure*}[t]
 \renewcommand{\captionlabelfont}{\footnotesize}
  \renewcommand\tabcolsep{2pt}
    \begin{center}
    \begin{tabular}{  c  c  c  c  c }
    \hline
    Original Image & Ours & DETR~\cite{carion2020end}  &MaskFormer~\cite{cheng2021per} & Ground Truth \\ 
    \hline
    & 50.6\% PQ & 45.1\% PQ & 47.6\% PQ& \\
    
    \begin{minipage}[b]{0.19\linewidth}
		\centering
	        {\includegraphics[width=\linewidth,height=3\linewidth]{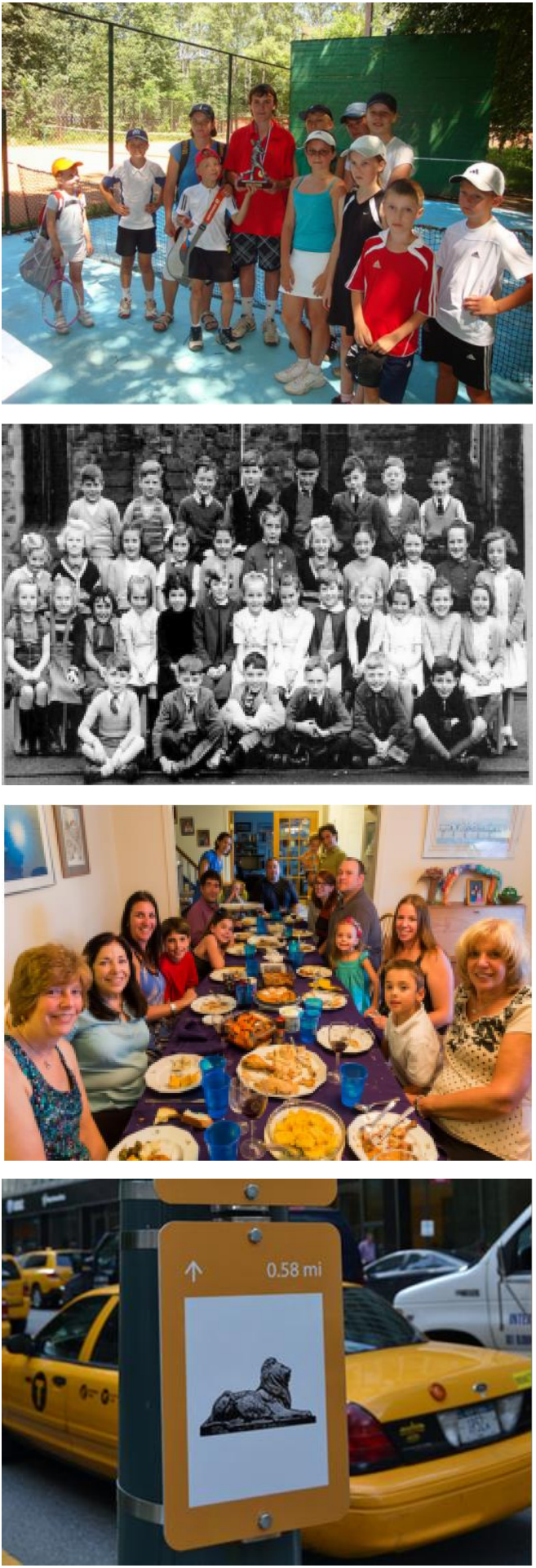}}
	        \end{minipage}
    & 
    \begin{minipage}[b]{0.19\linewidth}
		\centering
		{\includegraphics[width=\linewidth,height=3\linewidth]{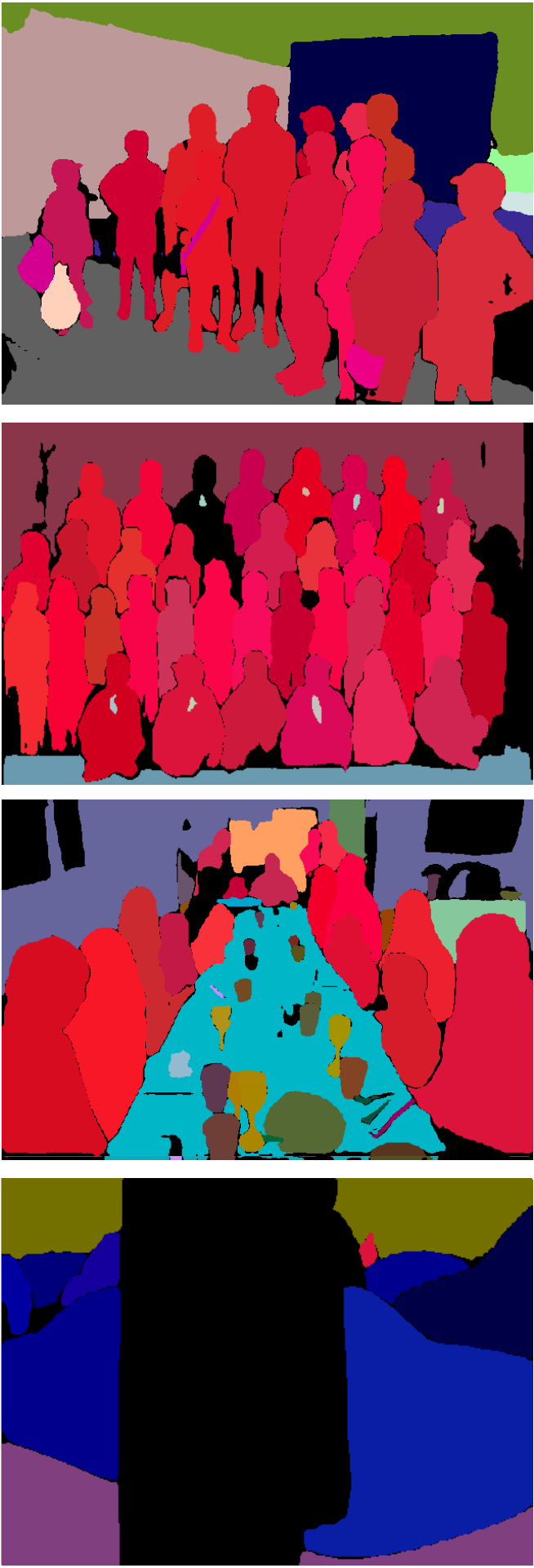}}
	\end{minipage}
	
    & \begin{minipage}[b]{0.19\linewidth} 
		\centering
		{\includegraphics[width=\linewidth,height=3\linewidth]{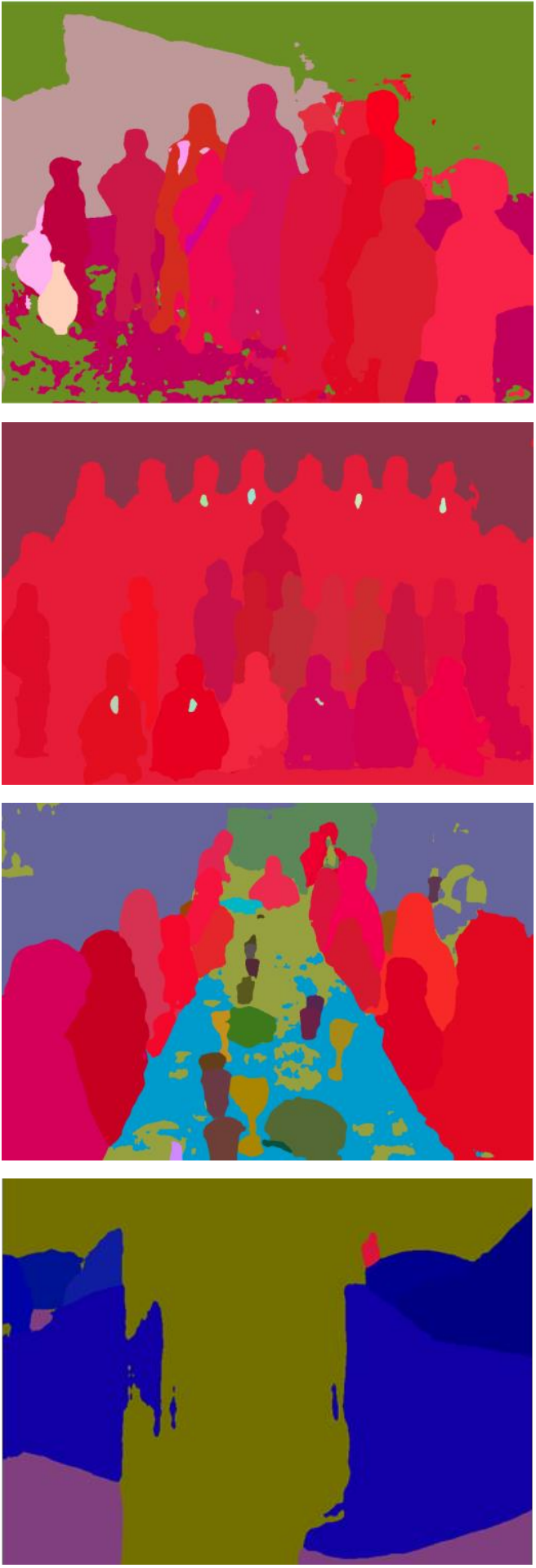}}
	\end{minipage}
	&
	\begin{minipage}[b]{0.19\linewidth}
		\centering
		{\includegraphics[width=\linewidth,height=3\linewidth]{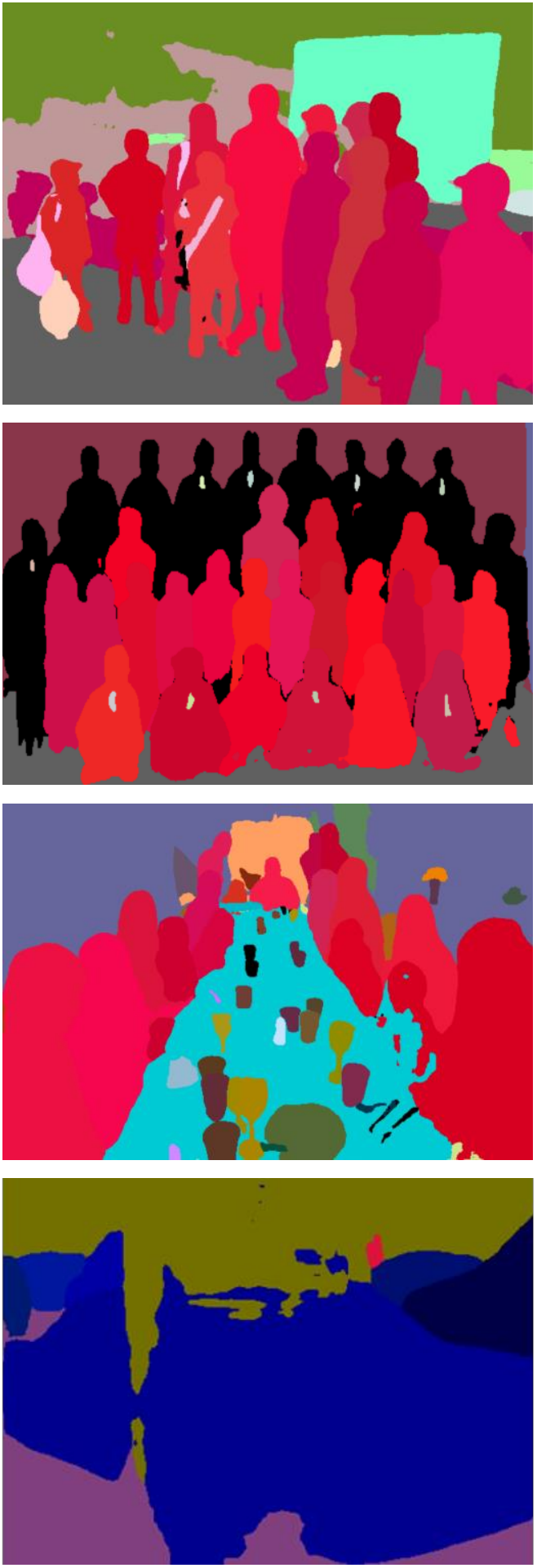}}
	\end{minipage}
	&
	\begin{minipage}[b]{0.19\linewidth}
		\centering
		{\includegraphics[width=\linewidth,height=3\linewidth]{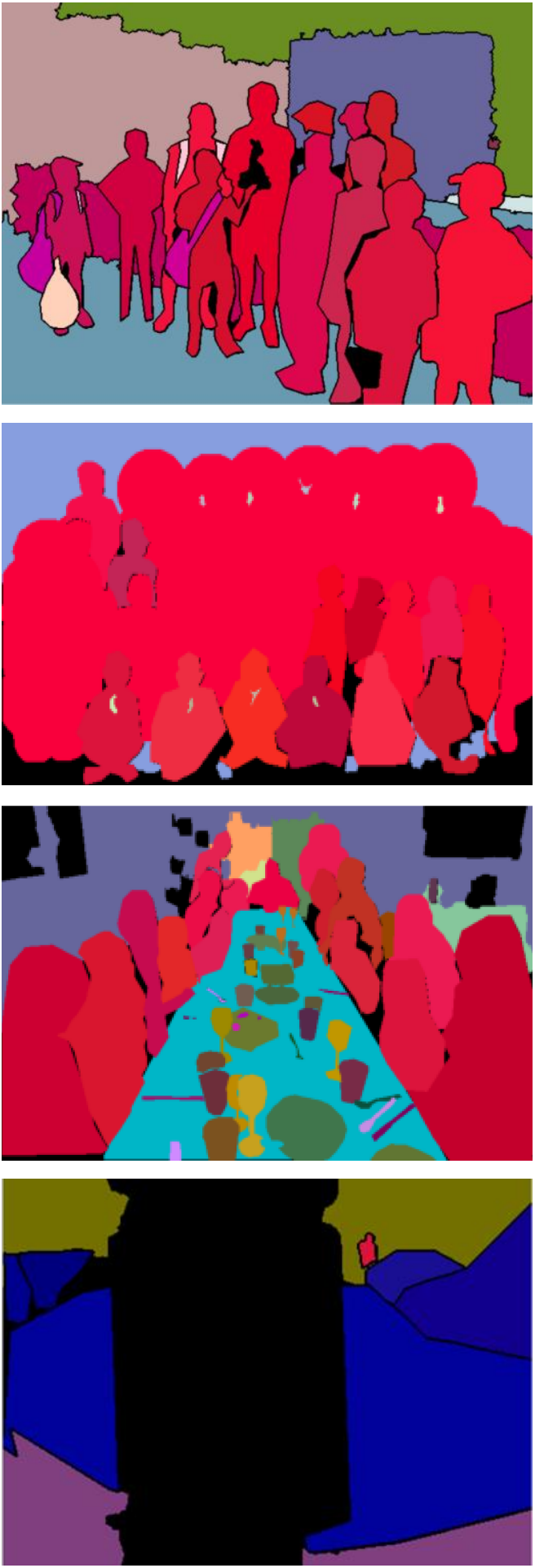}}
	\end{minipage}\\
  \end{tabular}
    \end{center}
    \vspace{-0.2in}
  \caption{ \footnotesize \textbf{Comparing visualization results of Panoptic SegFormer with other methods on the COCO \texttt{val} set.} For a fair comparison, all results are generated with ResNet-101~\cite{he2016deep} backbone. The second and fourth row results show that our method still performs well in highly crowded or occluded scenes. Benefits from our mask-wise inference strategy, our results have few artifacts, which often appear in the results of DETR~\cite{carion2020end} ({\eg, dining table of the third row}).}
  \label{visual}
  \vspace{-0.1in}
\end{figure*}

\begin{figure*}[t]
 \renewcommand{\captionlabelfont}{\footnotesize}
\begin{center}
   \includegraphics[width=0.95\linewidth,height=0.5\linewidth]{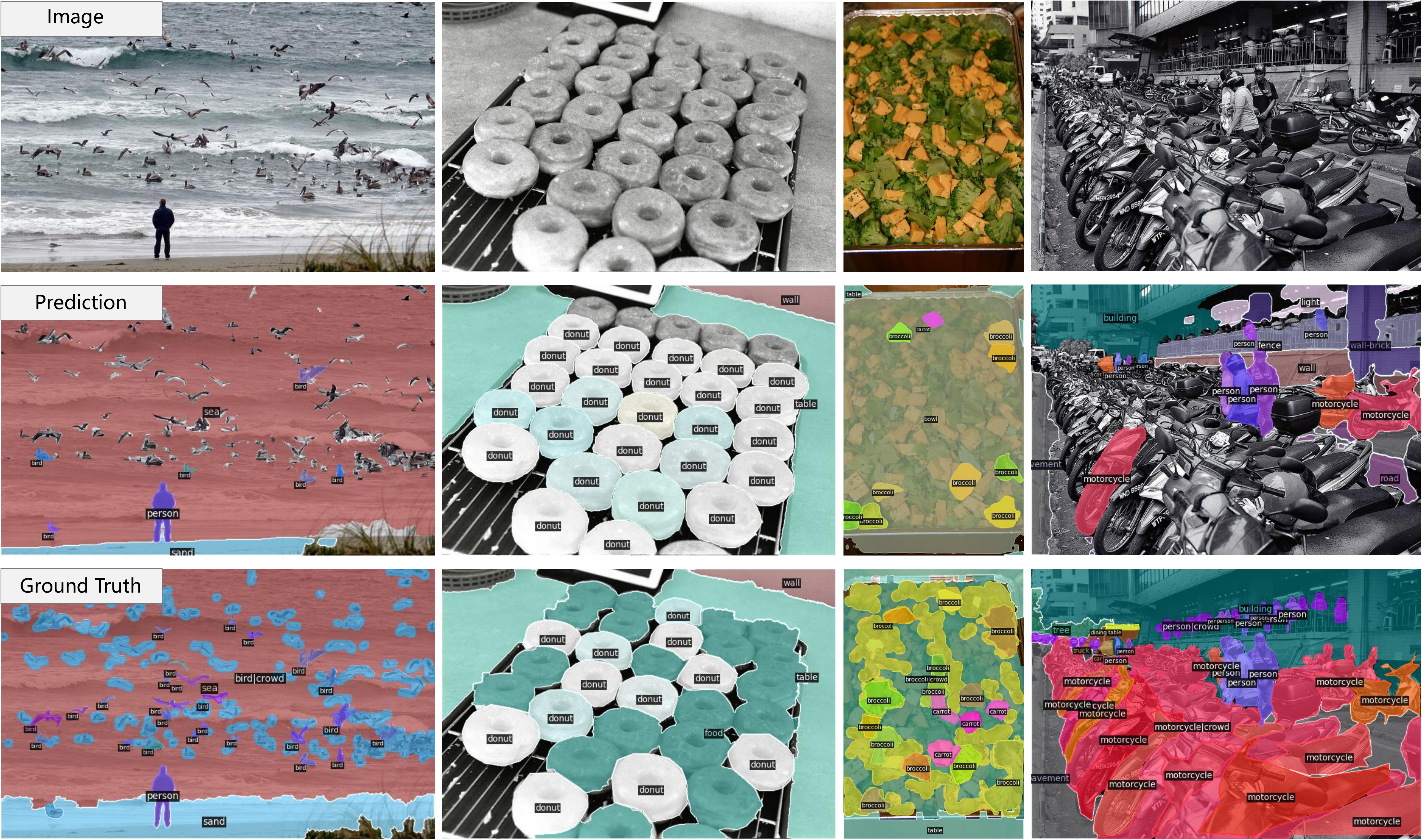}
\end{center}
\vspace{-0.2in}
   \caption{ \footnotesize 
    Failure case of Panoptic SegFormer.
   }
  \label{fail}
\end{figure*}

\begin{figure*}[t]
\begin{center}
   \includegraphics[width=0.95\linewidth]{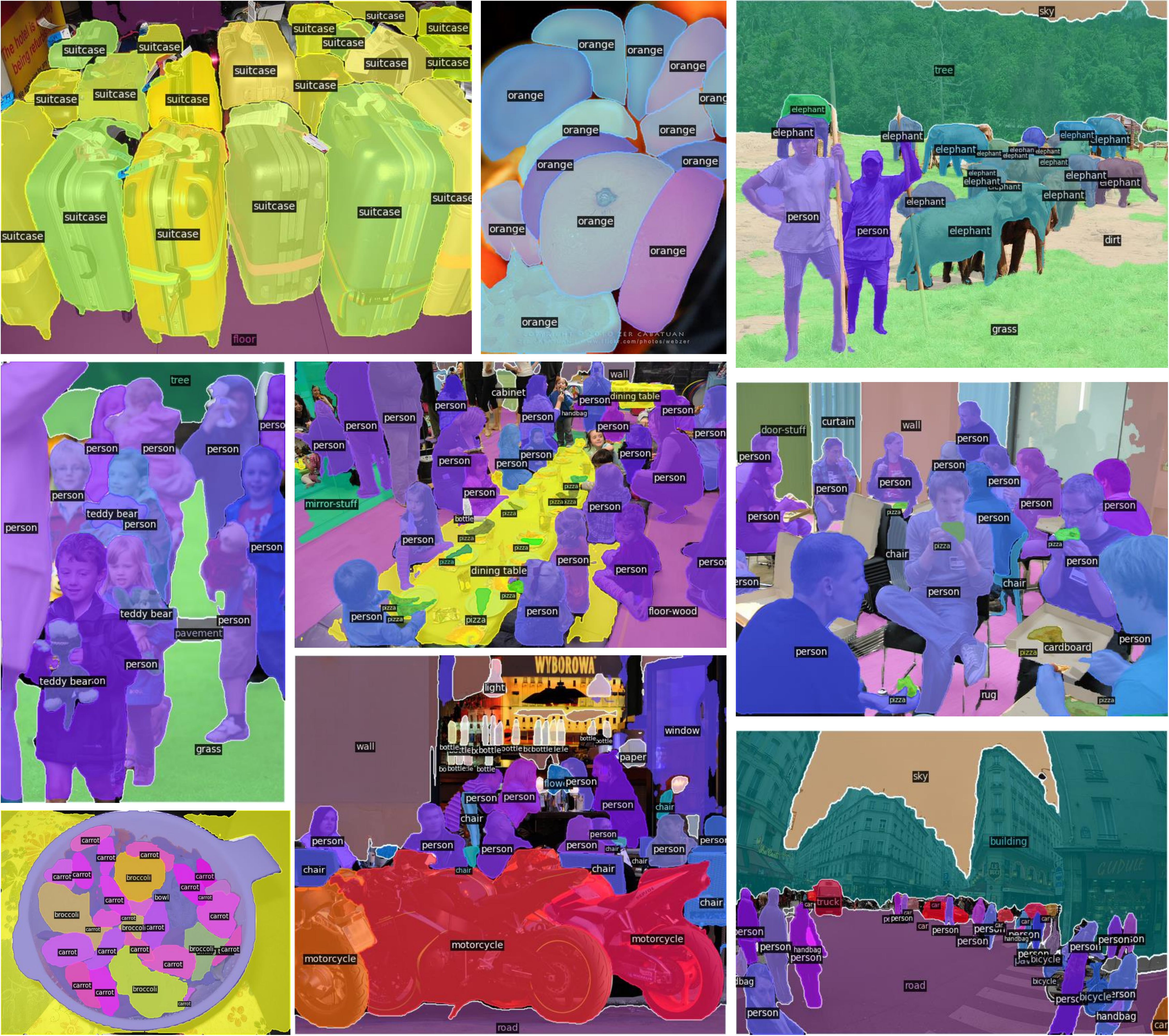}
\end{center}
\vspace{-0.2in}
   \caption{ \normalsize  Visualization results of some complex scenes.
   }
  \label{complex}
  \vspace{-0.1in}
\end{figure*} 

Although our proposed mask-wise merging strategy has achieved better results than other post-processing methods, it also has several shortcomings. First of all, we binaries the mask through a fixed threshold. This may cause one pixel to be easily assigned a void label because the values of all candidate instances at this pixel are below the threshold. Secondly, our strategy highly depends on the accuracy of confidence scores. If the confidence scores are not accurate, it will produce a low-quality panoptic format mask.

\subsection{Location Decoder}
Although we use the location decoder to detect the bounding boxes of things, our workflow is still very different from the previous box-based panoptic segmentation. For example, Panoptic FPN performs instance segmentation with Mask R-CNN style. The two-stage method usually needs to extract regions from the feature based on the bboxes and then use these regions to perform segmentation. The quality of segmentation is heavily dependent on the quality of detection. However, our location decoder is used to assist in learning the location clues of the query and distinguishing different instances. Mask will not have the wrong boundary due to the wrong boundary prediction of the bbox since the bbox does not constrain the mask. We also show that using mass centers of masks to replace bboxes can still learn location clues.

Another valuable function of the location decoder is to help filter out low-quality thing queries during the training and inference phase. This can greatly save memory.  Current transformer-based panoptic segmentation methods always consume a lot of GPU memory. For example,  MaskFormer takes up more than 20G of GPU memory with a batch size of 1 and R50 backbone.  Although these methods have achieved excellent results, they also require high hardware resources. However, our Panoptic SegFormer can be trained with taking up less than 12G memory by using a location decoder to filter out low-quality thing queries. In particular, we use bipartite matching for multiple rounds of matching in the detection phase. The thing query that already be matched will not participate in the next round of matching. After several rounds, we can select partial promising thing queries. Only these promising thing queries will be fed to the mask decoder. with this strategy, the mask decoder usually only needs to handle less than 100 thing and stuff queries.

\subsection{Mask Decoder}

\cref{panoptic_head} shows the architecture of DETR's panoptic head. Although it only contains 1.2M parameters, it has a huge computational burden (about $150$G FLOPs).
DETR adds ResNet features to each attention map, and this process repeats 100 times since there are 100 attention maps.
\cref{fig:mask_decoder_apx} shows the model architecture of our mask decoder. \cref{fig:mixer_apx} shows the process of converting multi-scale multi-head attention maps to mask.
We found that discarding the self-attention in the decoder does not affect the effectiveness of the model. The computational cost of our mask decoder is around 30G FLOPs.

\textbf{Multi-head attention maps.}
\cref{fig:mask} shows some samples of multi-head attention maps. Through a multi-head attention mechanism, different heads of one query learn their own attention preference. We observe that some heads pay attention to foreground regions, some prefer boundaries, and others prefer background regions. This shows that each mask is generated by considering various comprehensive information in the image. \cref{head} shows that utilizing a multi-head attention mechanism will outperform single-head attention by 0.4\% PQ.

\subsection{Advantage of Query Decoupling Strategy}\label{b.4}

DETR uses the same recipe to predict boxes of things and stuff
(To facilitate the distinction between the query decoupling strategy we proposed, we refer to the DETR's strategy as a joint matching strategy.). However, detecting bboxes for stuff like DETR is suboptimal. We counted the ratio of the area of masks to the area of bboxes on the COCO $\texttt{train2017}$. The ratios of things and stuff are 52.5\% and 9.2\%, separately. This shows that bounding boxes can not represent stuff well since the stuff is amorphous and dispersed. We also observe that bbox AP of DETR drops from 42.0 to 38.8 after training on panoptic segmentation. This may be due to the interference of stuff on things, since predicting stuff bboxes needs to re-adapt the model.

\cref{fig:detr_query} shows that DETR seems to learn automatic segregation between things and stuff, and each query either prefers things or stuff. However, we argue that this automatically learned segregation is not ideal. If one query prefers things, it will perform poorly when it generates stuff results. This situation is very common, and our following experiments based on Panoptic SegFormer will give detailed data.
Following DETR, we use 353 queries to predict things and stuff with the same recipe. Specifically, the input of the location decoder is 353 queries, which will detect both things and stuff. The refined queries are fed to the mask decoder to predict category labels and masks.
We define a query’s preference for things as P$_{t}$, which can be calculated by:

\begin{equation}\label{thing_preference}
{\rm P}_{t}^i = {\rm N}_{\rm things}^i/({\rm N}^i_{\rm things} + {\rm N}_{\rm stuff}^i),
\end{equation}
where ${\rm N}_{\rm things}$ and ${\rm N}_{\rm stuff}$ are the number of things and stuff masks that $i$-th query predicted on COCO val set.
${\rm P}_{t}^i>0.5 $ means that $i$-th query prefers things more than stuff. The predicted mask is a true positive (TP) if IoU between it and one ground truth mask is larger than 0.5 and the category of them is the same. Then we can calculate the precision of queries' predicted masks. 
\cref{query_apx} shows relevant statistical results.
First, we can observe that the queries that own lower P$_t$ basically have higher precision. The stuff precision of the queries that have the highest P$_{t}$ ($\left[0.9, 1.0\right]$) only is 0.30, which is much lower than the average stuff precision (0.60) on all queries. These erroneous results are mainly due to errors in the predicted category.
Queries that have no obvious preference for stuff and things( $P_t$ in $\left[0.4,0.6\right)$ ) performs poorly both on stuff and things.
These results demonstrate that using one query set to predict things and stuff simultaneously is flawed. This joint matching strategy is suboptimal for stuff.

In order to avoid mutual interference between stuff and things, we propose the query decoupling strategy to handle things and queries with a separate query set. Compared to stuff query, thing query will go through an additional location decoder.
However, all queries will produce the outputs in the same format. Things and stuff use the same loss for training, except that things use an additional detection loss. 
During inference, we can use our mask-wise merging strategy to merge them uniformly. This is different from the previous methods that modeled panoptic segmentation into instance segmentation and semantic segmentation. For example, Panoptic FPN uses one branch to generate things and one branch to generate stuff. The things and stuff generated by Panoptic FPN are in different formats and need different training strategies and post-processing methods. PQ$^{\rm st}$ with query decoupling outperforms joint matching strategy by 2.9\% PQ and experimental results verify the effectiveness of our method. The stuff precision by using query decoupling is 0.66, better than the joint matching strategy.

\section{Visualization}
\cref{visual} shows our visualization result against DETR and MaskFormer. We use the original codes that they officially implemented. First of all, compared with other methods, we can observe that our results are more consistent with ground truths. Due to the defects of pixel-wise argmax we discussed in \cref{armgax}, DETR always generates results with artifacts. MaskFormer performs better because they improved pixel-wise argmax by considering classification probabilities. However, it may still fail in hard cases. For example, it recognizes the billboard as a car in the fourth row. \cref{fail} shows some failure cases of our model. Firstly, our model may have lower recall when facing crowded scenarios filled with the same things, especially for the small targets. Another typical failure mode is that large stuff with a high confidence score occupies most of the space, causing other things not to be added to the canvas. \cref{complex} shows the results on some complex scenes.

\section{Various Backbones}
\begin{figure}[t]
\begin{center}
   \includegraphics[width=0.8\linewidth]{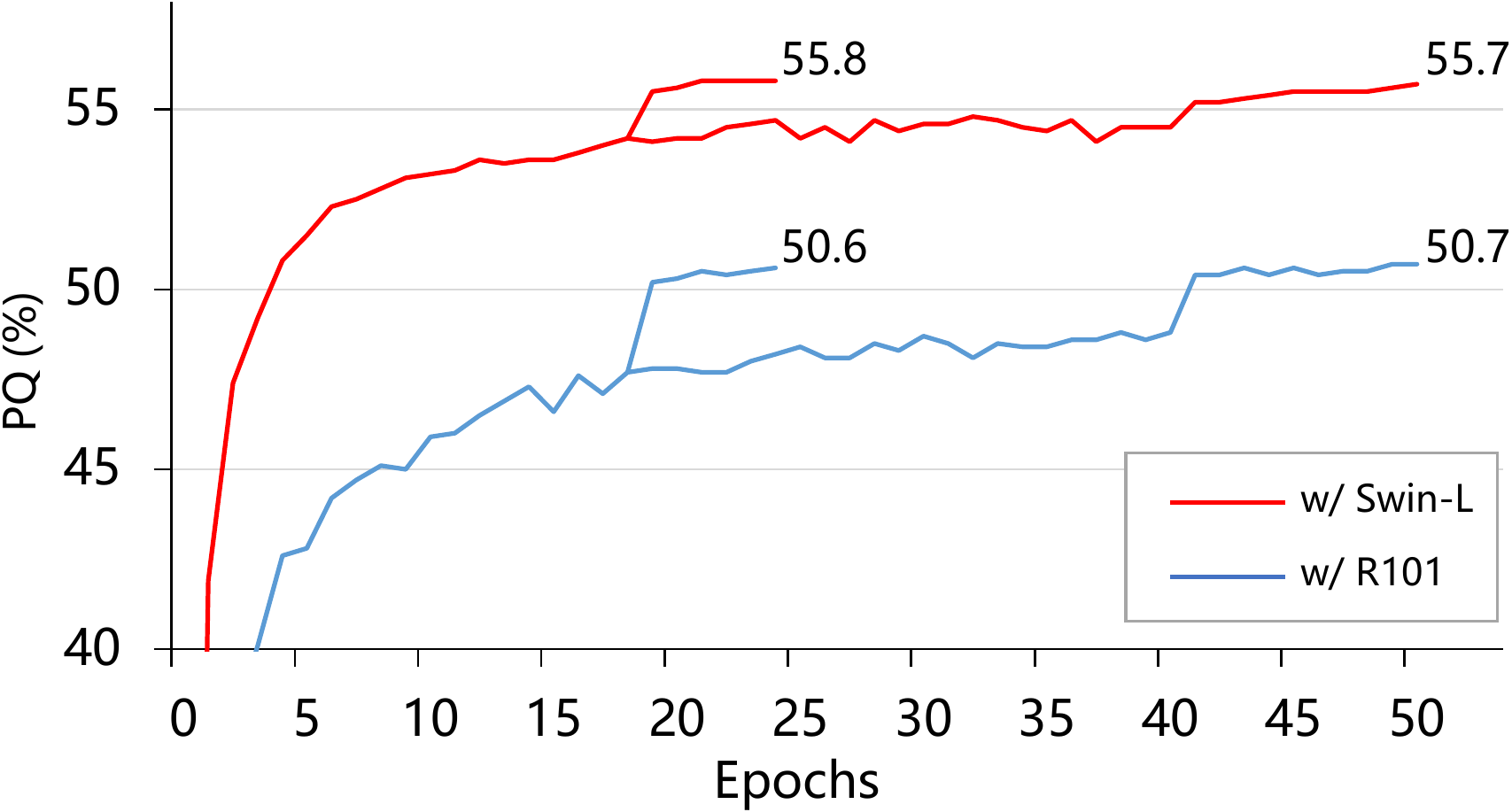}
\end{center}
\vspace{-0.2in}
   \caption{ \normalsize 
   By using ResNet-101~\cite{he2016deep} and Swin-L as the backbone, we train our model for 24 epochs and 50 epochs, separately. We can observe that our model that training for 24 epochs can achieve comparable or even higher results while comparing the models that training for 50 epochs.
   }
\label{fig:swin_r101}
\vspace{-0.1in}
\end{figure}

We give all the panoptic segmentation results under various backbones, as shown in \cref{backbone}. \cref{fig:swin_r101} shows two training curves with backbone ResNet-101 and Swin-L. With Swin-L, Panoptic SegFormer with training for 24 epochs even performs better than training for 50 epochs.
\begin{table*}
\begin{center}
\renewcommand\tabcolsep{10pt} 
\begin{tabular}{l c c c c c c c c c }
\toprule
Backbone & PQ & SQ & RQ &PQ$^{\rm th}$  & SQ$^{\rm th}$ & RQ$^{\rm th}$ & PQ$^{\rm st}$ &  SQ$^{\rm st}$ & RQ$^{\rm st}$ \\
\midrule

ResNet-50~\cite{he2016deep} & 49.6 & 81.6& 59.9 & 54.4&82.7 &65.1 & 42.4&79.9&52.1\\
ResNet-101~\cite{he2016deep} & 50.6& 81.9& 60.9& 55.5&83.0&66.3& 43.2&80.1&52.9\\
PVTv2-B0~\cite{wang2021pvtv2} & 49.5 & 82.4&59.2&55.3&83.3&65.8&40.6&80.9&49.2\\
PVTv2-B2~\cite{wang2021pvtv2} & 52.5&82.7&62.7&58.5&83.6&69.5& 43.4&81.4&52.4\\
PVTv2-B5~\cite{wang2021pvtv2} & 55.4&82.9&66.1&61.2&84.0&72.4&46.6&81.3&56.5\\
Swin-L~\cite{liu2021swin} & 55.8&82.6&66.8&61.7&83.7&73.3&46.9&80.9&57.0\\
\bottomrule
\end{tabular}
\end{center}
\vspace{-0.2in}
\caption{ \normalsize
Panoptic segmentation results on COCO val with various backbones.
}
\label{backbone}
\end{table*}

\section{Code and Data}
We use the official implementations of DETR\footnote{https://github.com/facebookresearch/detr}, MaskFormer\footnote{https://github.com/facebookresearch/MaskFormer}, Panoptic FCN\footnote{https://github.com/dvlab-research/PanopticFCN} to perform additional experiments. The models they provide all can reproduce the same scores they reported in their literature.
Deformable DETR is from Mmdet\footnote{https://github.com/open-mmlab/mmdetection}.

{\small
\bibliographystyle{unsrt}
\bibliography{egbib}
}

\end{document}